%% file: main.tex
\documentclass[10pt,letterpaper]{article}

\usepackage{xpeng-template}

\usepackage{hyperref}

\input{main/macros}

\xpengaffil{Robotics Foundation Model Team, Xpeng Inc.}

\title{IronViT: Toward Efficient Generalist Visual Representation Learning}

\author{%
  Jiaxi Huang$^{*}$, Yueqi Hu$^{*}$, Xin Zhu$^{*,\ddagger}$,
  Xiaopeng Zhang, Huiting Qiao, Yanglin Zhang, Zefeng Ji, Rongxue Li,
  Yifei Xu, Huiying Yu, Wei Liu, Jiayin Zheng, Yinggan Xu, Peipeng Chen,
  Yin Zhang$^{\S}$, Jian Yao$^{\S}$%
}

\date{}  

\begin{document}

\begingroup
\renewcommand{\thefootnote}{*}%
\footnotetext{Core contributors, in alphabetical order.}%
\renewcommand{\thefootnote}{\textdaggerdbl}%
\footnotetext{Project lead.}%
\renewcommand{\thefootnote}{\S}%
\footnotetext{Supervision.}%
\endgroup

\maketitle
\thispagestyle{titlepage}
\input{main/abstract}

\begin{center}
  \vspace{-0.25em}
  \includegraphics[width=0.82\textwidth]{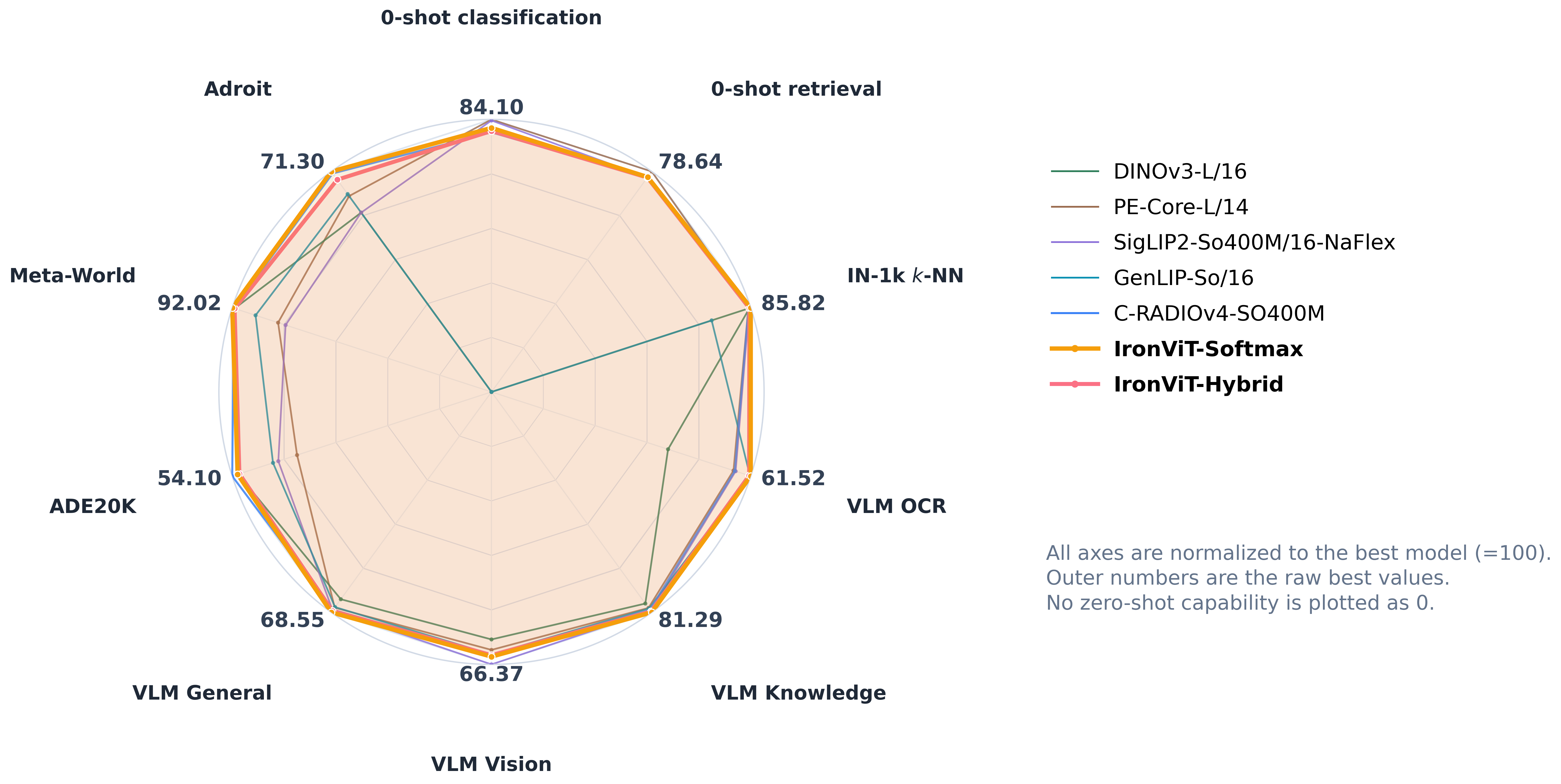}
  \captionsetup{font=small,skip=2pt,hypcap=false}
  \captionof{figure}{Capability profile of \methodname{} and representative baselines.}
  \label{fig:teaser-radar}
  \vspace{-0.5em}
\end{center}

\clearpage
\thispagestyle{fancy}
\tableofcontents
\clearpage

\input{main/introduction}
\input{main/data_curation}
\input{main/method}
\input{main/experiments}
\input{main/conclusion}

\bibliographystyle{unsrtnat}
\bibliography{refs}

\ifdefined\withoutappendix
\else
  \appendix
  \input{main/appendix}
\fi


\end{document}

%% file: main/macros.tex
\newcommand{\methodname}{IronViT\xspace}

\newcommand{\ironvitsoftmax}{\methodname-Softmax\xspace}
\newcommand{\ironvithybrid}{\methodname-Hybrid\xspace}

\providecommand{\todo}[1]{}
\renewcommand{\todo}[1]{\textcolor{red}{\textsf{[TODO: #1]}}}

\newcommand{\best}[1]{\textbf{#1}}
\newcommand{\second}[1]{\underline{#1}}
\newcommand{\pmstd}[1]{{\tiny\,$\pm$#1}}

\SetMathAlphabet{\mathcal}{normal}{OMS}{cmsy}{m}{n}
\SetMathAlphabet{\mathcal}{bold}{OMS}{cmsy}{b}{n}
\newcommand{\imganchors}{\ensuremath{\mathcal{A}^{\mathrm{img}}}}
\newcommand{\txtanchors}{\ensuremath{\mathcal{A}^{\mathrm{txt}}}}
\newcommand{\teachers}{\ensuremath{\mathcal{T}}}                 
\newcommand{\studadapter}[1]{\ensuremath{A_{#1}}}               

\ifdefined\withoutappendix
  \renewcommand{\appref}[1]{the full version}
\fi

%% file: main/abstract.tex
\begin{abstract}
A generalist vision encoder must capture semantic, spatial, language-aligned, and
action-relevant cues within a unified representation, yet softmax
attention underlying today's most capable visual backbones becomes
prohibitively expensive at high resolution. A natural attempt to address both
challenges is to distill multiple specialist teachers directly into an efficient
architecture. We find that directly coupling these objectives degrades representation quality,
as the student must simultaneously reconcile heterogeneous
capabilities and adapt them to a different token-mixing architecture.
We introduce \methodname{}, built on a
simple principle: \emph{consolidate capabilities before constraining
computation}. \methodname{} first distills complementary specialists into a
softmax attention capability bridge, then progressively transfers the consolidated
representation to a hybrid softmax--linear attention encoder. A purpose-built data pipeline further curates the distillation corpus
for higher information density and broader domain coverage. Across recognition, retrieval,
dense prediction, multimodal understanding, and robotic learning,
\methodname{} is competitive with leading specialist and generalist vision
encoders. The softmax bridge achieves the strongest aggregate performance
in multimodal understanding and robotic learning among the evaluated backbones, while the hybrid
encoder retains broad transfer performance with an efficiency advantage that grows
with input resolution. Together, these results show that consolidating
capabilities before architectural conversion can yield a generalist visual
encoder without inheriting the prohibitive high-resolution cost of
conventional softmax attention.
\end{abstract}

%% file: main/introduction.tex
\section{Introduction}
\label{sec:intro}

Visual representation learning is shifting from task-specific models toward
general-purpose backbones that support a wide range of perception and embodied
intelligence tasks. Modern Vision Transformers (ViTs) now transfer across
image classification~\cite{dosovitskiy2021vit}, semantic and instance
segmentation~\cite{oquab2024dinov2,kirillov2023sam}, monocular depth
estimation~\cite{yang2024depth}, vision--language
understanding~\cite{zhai2023siglip}, and robotic
control~\cite{kim2024openvla}.
Yet progress in these areas has been driven by different supervision sources and
optimization objectives, resulting in representations with distinct strengths
and trade-offs. Classification models emphasize category-level semantics but
often underrepresent fine-grained spatial structure; dense prediction models
preserve local detail but are less suited to open-vocabulary recognition;
vision--language models align visual and linguistic spaces but may sacrifice
dense-feature quality; and robot-learning representations must capture
action-relevant cues that are rarely emphasized in static image datasets.
Obtaining a backbone that performs strongly across all of these capabilities is
therefore difficult under any single training paradigm.

Multi-teacher distillation~\cite{ranzinger2024amradio,heinrich2024radiov25,zhu2026eupe,shang2024theia} provides a natural
way to bridge this fragmentation.
Rather than learning semantic, geometric, language-aligned, and
action-relevant representations entirely from raw data, a student can inherit
capabilities already encoded by specialized foundation models. This is
particularly attractive when training data and compute are limited, as
distillation can directly exploit mature teacher feature spaces instead of
rediscovering these representations from scratch. Consolidating heterogeneous
teachers, however, remains challenging. Their representations differ in
dimensionality, spatial resolution, semantic abstraction, and invariance
properties, while their supervisory signals may be redundant or conflicting on
the same input. A successful generalist therefore requires a coherent
representation space in which complementary capabilities can coexist.

Beyond capability consolidation, efficiently deploying a generalist vision
backbone remains challenging. Most high-capacity vision foundation models rely
on softmax self-attention, whose time and memory costs grow quadratically with
the number of visual tokens. This bottleneck becomes particularly pronounced in
dense prediction, high-resolution vision--language reasoning, multi-view
perception, and temporally extended robotic observations, where preserving
spatial or temporal detail produces long token sequences. Linear attention
mechanisms~\cite{katharopoulos2020transformers} can reduce this quadratic
dependence, but they exhibit optimization behavior and inductive biases that
differ from those of softmax attention. Training such architectures typically
requires substantial pretraining, while directly replacing softmax attention in
a pretrained model can disrupt representations that are important for
downstream transfer.

A seemingly natural solution is to distill all specialist teachers directly
into a hybrid attention student. This formulation, however, couples two
distinct transfer problems. The first is a \textbf{capability gap}: the student
must reconcile heterogeneous semantic, spatial, geometric, multimodal, and
action-relevant capabilities. The second is an \textbf{architectural gap}:
representations learned with softmax attention must be transferred to a
different token-mixing architecture. Solving both simultaneously requires the
hybrid attention student to discover a shared representation space while
adapting that space to a new attention operator. Architectural mismatch may
interfere with teacher aggregation, while conflicting teacher signals may in
turn hinder architectural adaptation. This motivates a simple principle:
\textbf{Capabilities should be consolidated before computation is constrained.}

Based on this principle, we introduce \methodname, a two-stage framework for
building an efficient generalist vision backbone. We first distill complementary
specialist teachers into a unified softmax attention ViT whose representations
cover recognition, vision--language, and robotic tasks. This model serves as an
intermediate capability bridge that consolidates heterogeneous knowledge into a
coherent representation. We then distill the unified model into a hybrid
attention backbone, transferring its learned capabilities while improving
efficiency at high resolution and long sequence lengths. By separating
capability consolidation from architectural adaptation, our framework
decomposes the joint multi-teacher, cross-architecture problem into two more
controlled stages.

This decoupling also places stronger demands on the training data.
Multi-teacher distillation benefits from samples that provide informative
supervision across complementary capabilities while limiting redundancy under a
practical training budget. We therefore construct a purpose-built data pipeline
that selects, enriches, and balances training samples for multi-teacher
distillation. With this pipeline, \methodname{} achieves broad multi-task
transfer using fewer than 100M curated training images and a total of 2,336 H200
GPU-hours across both distillation stages.

The benchmark-level profile in \figref{fig:teaser-radar} summarizes the
empirical performance of \methodname{} across representative recognition,
retrieval, dense prediction, multimodal reasoning, and robotic-control
evaluations. The intermediate softmax attention model successfully consolidates
complementary teacher capabilities, while the resulting hybrid attention model
preserves most of this performance with substantially improved efficiency in
high-resolution and long-sequence regimes. Moreover, the two-stage formulation
consistently outperforms direct multi-teacher distillation into the same hybrid
attention student across our ablation settings, supporting the value of
capability consolidation before architectural adaptation.

The remainder of this paper is organized as follows. \secref{sec:data}
details our distillation-oriented data curation pipeline, including
redundancy reduction, coverage-aware selection, and target-domain
enrichment. \secref{sec:method} formulates our two-stage framework,
from heterogeneous capability consolidation in a softmax attention model
to cross-architecture transfer into a hybrid attention backbone.
\secref{sec:experiments} evaluates \methodname{} across classification,
image--text retrieval, dense prediction, vision--language understanding,
and robotic learning, and further examines its accuracy--efficiency
trade-off and the contribution of each design component.

%% file: main/data_curation.tex
\section{Data Curation}
\label{sec:data}

Large-scale image collections can contain substantial low-quality and
redundant data, while providing uneven coverage of visual concepts and target
domains. Motivated by data-centric pretraining~\cite{gadre2023datacomp}, we
develop a scalable, modular image-curation framework to address these
complementary challenges and construct visual training corpora tailored to our
multi-stage training procedure.

Starting from a noisy web-scale image pool, the framework removes invalid and
low-quality samples, suppresses both low-level and semantic redundancy, rebalances 
the distribution of visual concepts, and enriches underrepresented target domains with
additional relevant images. It is designed to improve the information density
and semantic coverage of the training data, supporting more efficient visual
representation learning.

\begin{figure}[t]
  \centering
  \includegraphics[width=\linewidth]{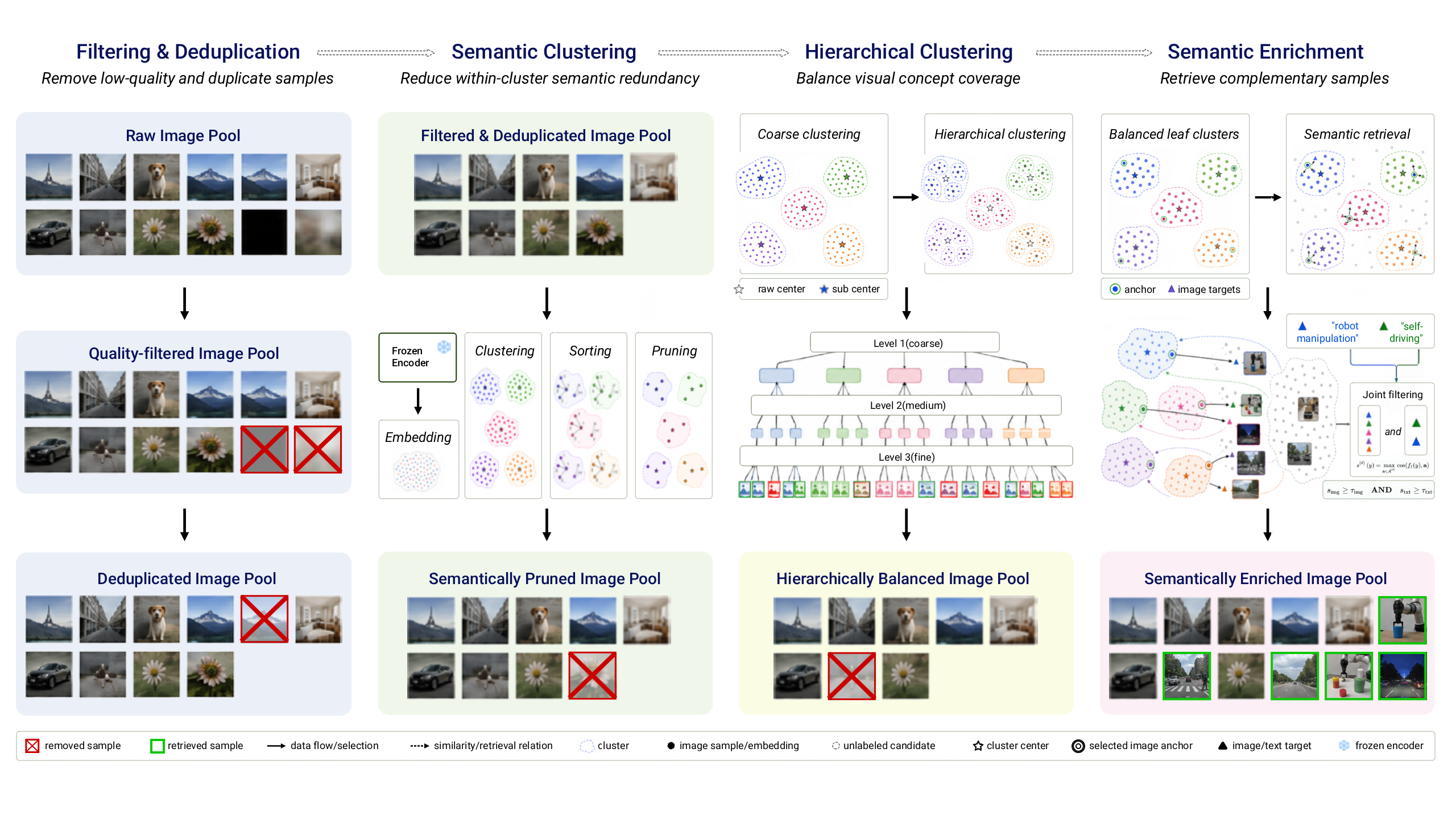}
  \caption{Overview of the proposed data curation framework.}
  \label{fig:data_pipeline}
\end{figure}

As illustrated in \figref{fig:data_pipeline}, the framework comprises four
modules: image-quality filtering and perceptual deduplication; semantic
clustering with in-cluster pruning, following
SemDeDup~\cite{abbas2023semdedup}; hierarchical $k$-means with resampling and
balanced sampling~\cite{vo2024autocuration}; and target-aware semantic
enrichment inspired by SSE~\cite{shen2025sse}. These modules are applied 
selectively to construct the pretraining and refinement corpora, as detailed in
\secref{sec:data-recipe}.

\subsection{Filtering and deduplication}
\label{sec:data-filter}

\paragraph{Quality filtering.}
Given the raw image corpus $\mathcal{D}_{raw}$, We apply conservative image-level 
checks to remove invalid or severely degraded inputs before subsequent embedding-based 
processing. These checks use only image content and metadata and cover three categories:
\begin{enumerate}
  \item \textbf{Decoding and format failures}: images that cannot be decoded,
    have unsupported formats, corrupted byte streams, or inconsistent image
    metadata are removed.

  \item \textbf{Resolution and geometry outliers}: images with extremely small
    or large resolutions, invalid dimensions, or extreme aspect ratios are
    filtered to stabilize the visual input distribution.

  \item \textbf{Low visual quality}: images with severe blur or visual
    degradation are discarded using conservative quality heuristics.
\end{enumerate}
These checks target basic image validity and usability rather than semantic
relevance.

\paragraph{Perceptual deduplication.}
We use DCT-based perceptual hashing~\cite{zauner2010phash} to identify low-level visual duplicates and
remove redundant copies. This procedure targets near-identical image
variants rather than semantically similar but visually distinct samples;
the latter are addressed by the embedding-based pruning module in
\secref{sec:data-semcluster}.

\subsection{Semantic clustering and in-cluster pruning}
\label{sec:data-semcluster}

Perceptual deduplication does not fully address redundancy in visual content.
Following SemDeDup~\cite{abbas2023semdedup}, we identify semantically
redundant images in a pretrained embedding space and restrict similarity
comparisons to local clusters, avoiding exhaustive comparisons over the
entire corpus.

\paragraph{Image embedding.}
We extract image features using pretrained CLIP-style
encoders~\cite{radford2021clip}, specifically ViT-H/14
models loaded through OpenCLIP. The encoders remain frozen, and each
embedding $\mathbf{e}_i \in \mathbb{R}^D$ is $L_2$-normalized before
clustering and similarity computation.

\paragraph{Semantic clustering.}
We partition the embeddings into $K$ clusters using spherical $k$-means
implemented in FAISS~\cite{johnson2019faiss}. Each image is assigned to
its most similar centroid:
\[
c_i = \arg\max_{c \in \{1,\ldots,K\}}
      \mathbf{e}_i^\top \boldsymbol{\mu}_c,
\qquad
d_i = 1-\mathbf{e}_i^\top \boldsymbol{\mu}_{c_i},
\]
where $\boldsymbol{\mu}_c$ denotes the unit-normalized centroid of
cluster $c$, and $d_i$ is the sample's distance to its assigned centroid.
The resulting clusters define the neighborhoods used for redundancy
detection.

\paragraph{In-cluster semantic pruning.}
Following SemDeDup, samples within each cluster are sorted by decreasing
centroid distance, giving priority to less central samples. For a cluster
of size $m$, let $\mathbf{e}_{(1)},\ldots,\mathbf{e}_{(m)}$ denote the
embeddings in this order. A sample at position $j>1$ is removed if
\begin{equation}
  \max_{1 \leq i < j}
  \mathbf{e}_{(i)}^\top \mathbf{e}_{(j)} > 1-\varepsilon,
  \label{eq:semantic-pruning}
\end{equation}
where all preceding samples participate in the comparison, regardless of
whether they are themselves marked for removal. The first sample in each
cluster is retained. We use a cosine-distance threshold of
$\varepsilon=0.15$, corresponding to a cosine-similarity threshold of
$0.85$.

Our implementation computes similarities blockwise to avoid materializing
the full within-cluster similarity matrix while preserving the pruning
criterion. The retained images form the semantically pruned pool used
for subsequent hierarchical balanced sampling.

\begin{figure}[t]
    \centering
    \includegraphics[width=\linewidth]{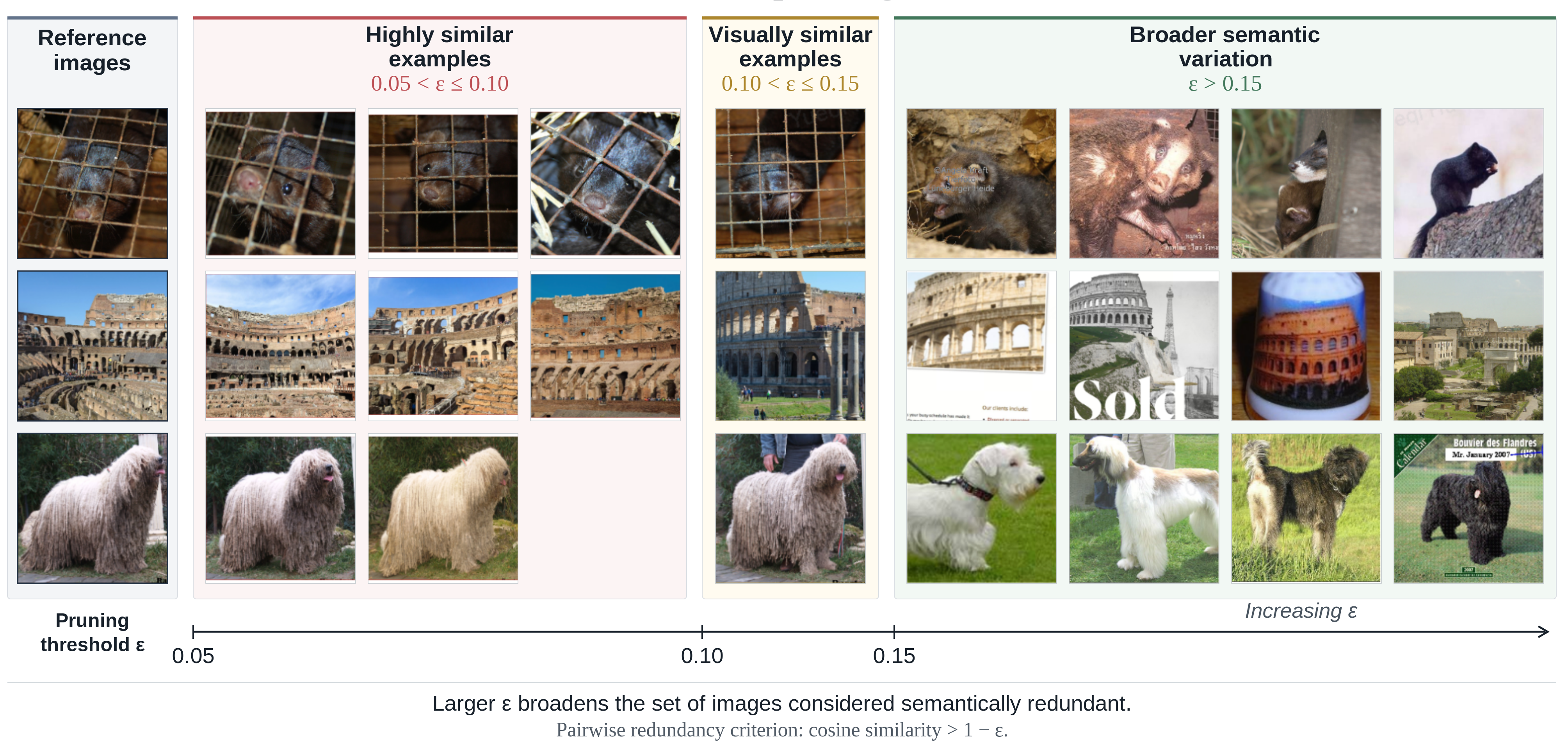}
    \caption{
    \textbf{Effect of the semantic pruning threshold $\varepsilon$.}
    Increasing $\varepsilon$ broadens the neighborhood considered semantically redundant, 
    resulting in progressively more aggressive in-cluster pruning.
    }
    \label{fig:epsilon_pruning}
\end{figure}

\subsection{Hierarchical balanced sampling}
\label{sec:data-hkm}

Semantic pruning reduces local redundancy but does not necessarily balance
the distribution of visual concepts. We therefore adopt hierarchical
balanced sampling following Vo et al.~\cite{vo2024autocuration} to select
a fixed-size subset while mitigating the overrepresentation of frequent
visual concepts.

\paragraph{Hierarchical clustering with resampling.}
We build a bottom-up clustering hierarchy over the retained image embeddings.
The first level partitions image embeddings into fine-grained clusters,
while each subsequent level clusters the centroids from the preceding
level. Following~\cite{vo2024autocuration}, we refine the hierarchy through
iterative resampling: up to $r_\ell$ centroid-nearest input points are
selected from each cluster at level $\ell$, and the resulting subset is
used to recompute the centroids. Reassignment and resampling are repeated
to reduce the influence of the original sample density on centroid
allocation.

\paragraph{Top-down balanced sampling.}
Given a target subset size $N_{\mathrm{target}}$, we recursively allocate
the sampling budget from the coarsest clusters to the finest ones.
For a parent node $p$, let $\mathcal{J}_p$ denote its children,
$B_p$ its assigned budget, and $m_j$ the number of available images
in the subtree rooted at child $j$. Initial child quotas are
\begin{equation}
  \widetilde{n}_j = \min(q_p,m_j),
  \qquad
  \sum_{j \in \mathcal{J}_p}\widetilde{n}_j \leq B_p,
  \label{eq:waterfill}
\end{equation}
where $q_p$ is the largest integer not exceeding
$\max_{j \in \mathcal{J}_p}m_j$ that satisfies the budget constraint.
Any remaining budget is distributed by assigning one additional sample
to randomly chosen children with unused capacity, yielding integer
quotas that sum exactly to $B_p$.

This allocation is applied recursively, starting with
$B_{\mathrm{root}}=N_{\mathrm{target}}$. At the finest level, images are
sampled uniformly at random without replacement. The procedure balances
budgets among sibling subtrees while respecting their capacities,
rather than enforcing identical sample counts across all clusters.

\paragraph{Implementation adaptations.}
We reuse the normalized CLIP embeddings from
\secref{sec:data-semcluster} and implement spherical $k$-means with
FAISS~\cite{johnson2019faiss}, replacing the Euclidean clustering
used in the reference implementation. To limit the cost of finest-level
clustering, centroid fitting and resampling operate on a fixed subset
of the retained embeddings. The final centroids are then used to assign
the full retained pool in chunks; subsequent sampling budgets are
computed from these full-pool assignments, not from subsample counts.
Higher levels operate on the complete centroid set from the preceding
level. This subsampling approximates hierarchy construction while
retaining the capacity-constrained sampling procedure described above.

\subsection{Semantic enrichment}
\label{sec:data-enrichment}

Generic image collections may provide insufficient coverage of target visual
domains. Inspired by the semantic enrichment framework of
SSE~\cite{shen2025sse}, we retrieve relevant images from an external
unlabeled pool to supplement these domains. Whereas SSE uses generated
caption embeddings to expand semantic coverage, our approach performs
target-conditioned retrieval using image and text anchors in a shared CLIP
embedding space. The external pool aggregates multiple public and in-house
image collections, including web-scale sources such as
DataComp-1B~\cite{gadre2023datacomp}.

\paragraph{Target anchors.}
Given a target-domain seed image set, we apply spherical $k$-means to its
normalized CLIP embeddings and select the embedding of the nearest seed
image to each centroid as an image anchor. This yields a compact anchor
set $\imganchors$ representing the target visual
distribution. We additionally encode domain-level text prompts using
the corresponding text encoder to obtain
$\txtanchors$. These prompts describe the desired domains
rather than individual seed images. All image and text embeddings are
$L_2$-normalized.

\paragraph{Joint image--text selection.}
For each candidate image with embedding $\mathbf{e}_r$, we compute its
maximum similarity to each anchor set:
\begin{align}
  s_{\mathrm{img}}(r)
  &= \max_{\mathbf{a}\in\imganchors}
     \mathbf{e}_r^\top\mathbf{a},
  \label{eq:img-sim} \\
  s_{\mathrm{txt}}(r)
  &= \max_{\mathbf{t}\in\txtanchors}
     \mathbf{e}_r^\top\mathbf{t}.
  \label{eq:txt-sim}
\end{align}
Because the embeddings are normalized, these inner products equal cosine
similarities. A candidate is selected only when
\begin{equation}
  s_{\mathrm{img}}(r)\geq\tau_{\mathrm{img}}
  \quad\land\quad
  s_{\mathrm{txt}}(r)\geq\tau_{\mathrm{txt}}.
  \label{eq:joint-enrichment}
\end{equation}
The image constraint measures similarity to the seed examples, while the
text constraint provides complementary domain relevance. The thresholds
are calibrated separately using the empirical score percentiles, without
assuming that image--image and image--text similarities share the same
numerical scale.

\subsection{Training corpora}
\label{sec:data-recipe}

We construct two image corpora using the curation modules described above:
a pretraining corpus $\mathcal{D}_{\mathrm{pre}}$ and a refinement
corpus $\mathcal{D}_{\mathrm{ref}}$. Their use in the multi-stage
training procedure is described in \secref{sec:method}.

\begin{figure}[t]
  \centering
  \includegraphics[width=\linewidth]{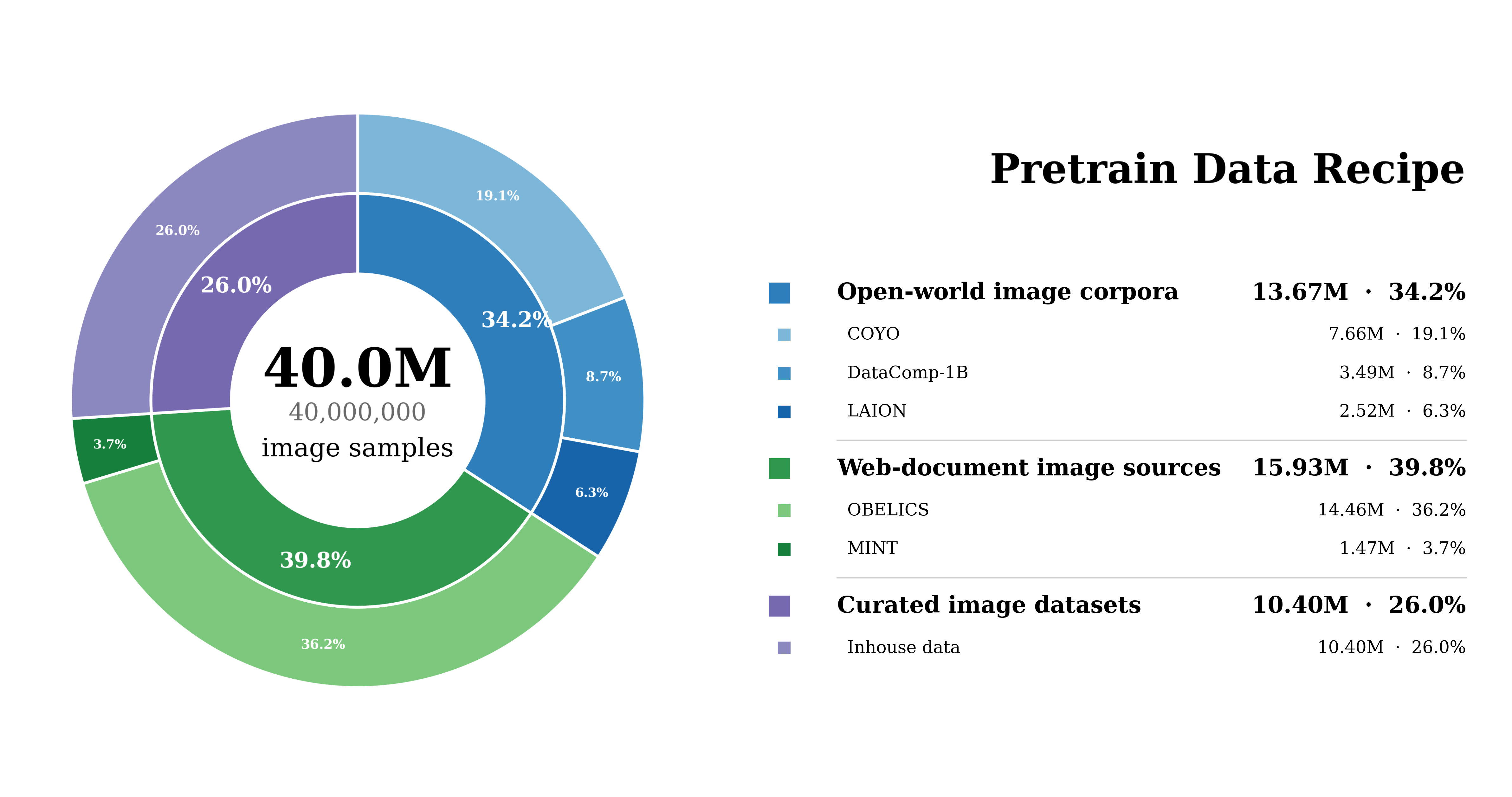}
  \caption{
  \textbf{Recipe for the pretraining candidate pool.}
  The inner and outer rings show recipe components and their constituent
  sources, respectively. Counts and percentages refer to the 40.0M-image
  candidate pool after curation.
  }
  \label{fig:pretrain_recipe}
\end{figure}

\paragraph{Pretraining corpus.}
We assemble a candidate pool of approximately 85.5 million image samples
from open-world image corpora, web-document sources, and curated image
collections. Image filtering and perceptual deduplication
reduce the pool to approximately 79 million samples, and semantic
clustering with in-cluster pruning further reduces it to approximately
64 million. Hierarchical balanced sampling then selects 40 million
samples to form our final $\mathcal{D}_{\mathrm{pre}}$. Its source composition 
is summarized in \figref{fig:pretrain_recipe}. 

\begin{figure}[t]
  \centering
  \includegraphics[width=\linewidth]{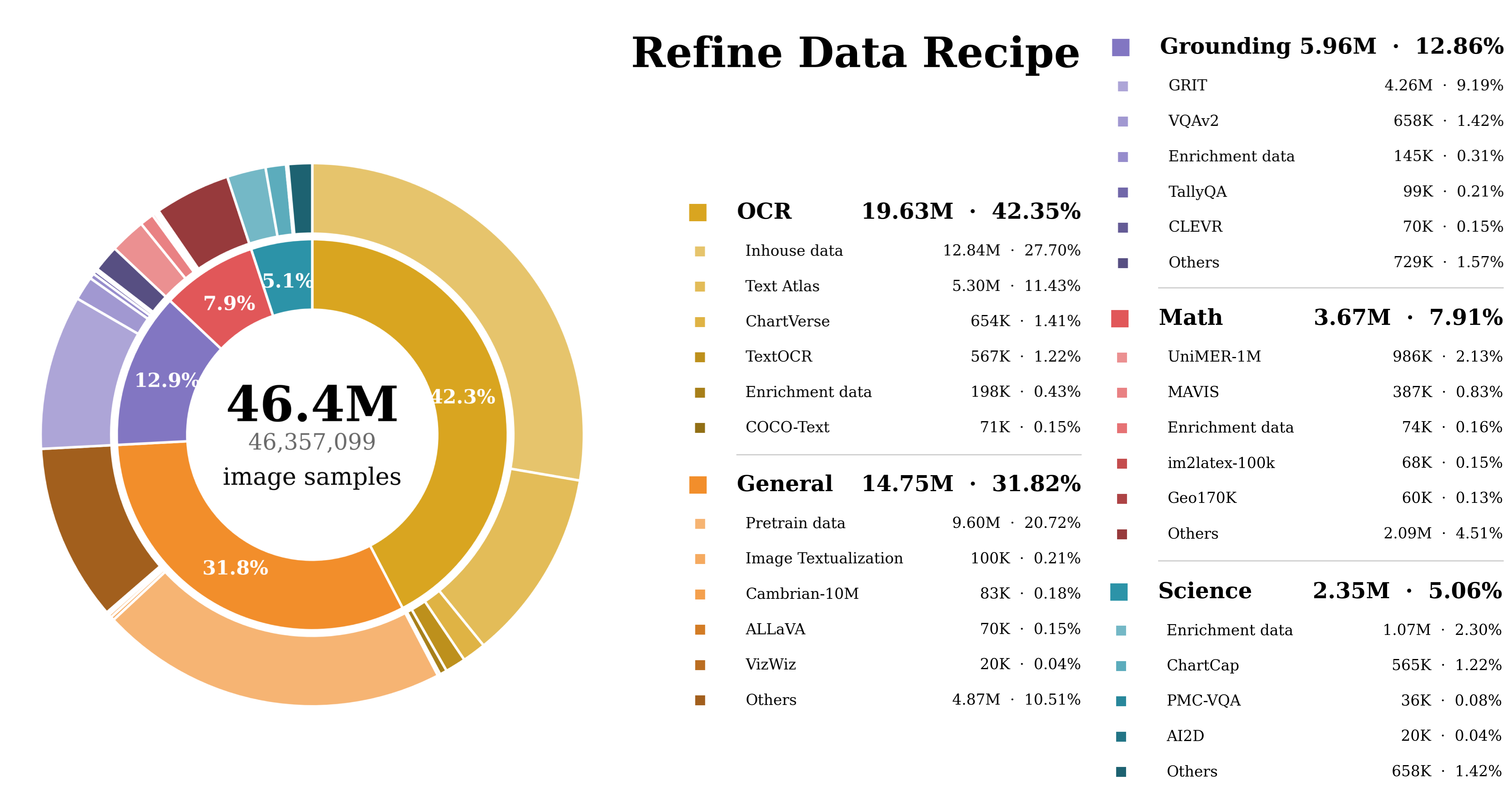}
  \caption{
  \textbf{Recipe for the refinement corpus after semantic enrichment.}
  The inner and outer rings show recipe components and their constituent
  sources, respectively. Counts and percentages refer to the final 46.4M-image
  refinement corpus.
  }
  \label{fig:refine_recipe}
\end{figure}

\paragraph{Refinement corpus.}
The refinement corpus $\mathcal{D}_{\mathrm{ref}}$ combines
capability-oriented image collections, including OCR, grounding, math,
and science, with a subset of the pretraining data. Following
image-level filtering, we apply semantic enrichment to supplement
target visual domains. The final corpus contains approximately
46.4 million image samples, including 1.487 million samples retrieved
through semantic enrichment. Its post-enrichment composition is
summarized in \figref{fig:refine_recipe}.

The cumulative ablation in \secref{sec:exp-abl-data}
(\figref{fig:abl-data}) evaluates the contributions of image filtering,
semantic pruning, and hierarchical balanced sampling to representation
quality.

%% file: main/method.tex
\section{Method}
\label{sec:method}

\begin{figure*}[t]
  \centering
  \includegraphics[width=\linewidth]{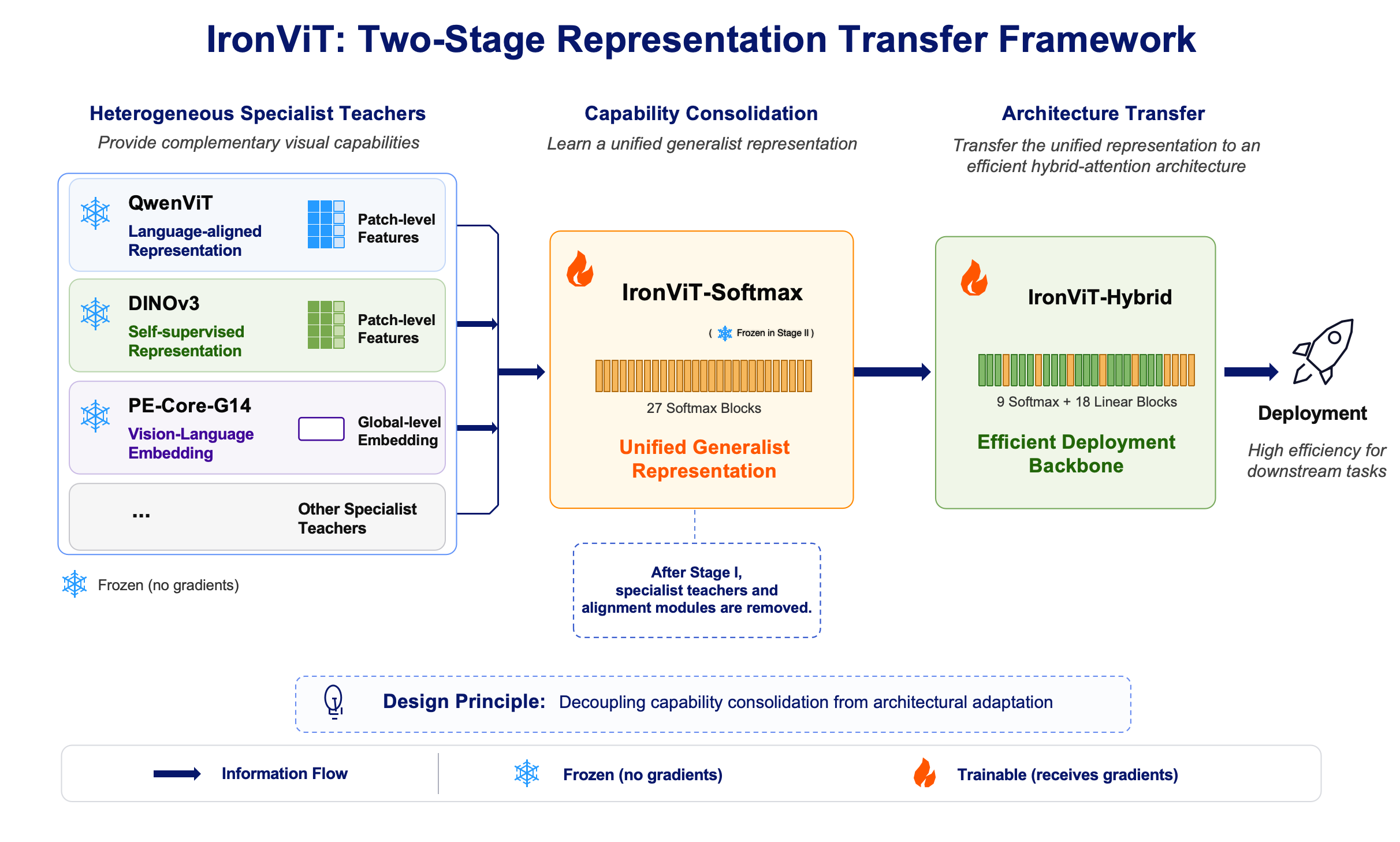}
  \caption{Overview of \methodname{}. Stage~I consolidates complementary
  specialist capabilities into a softmax attention bridge, and Stage~II transfers
  the unified representation to a hybrid attention encoder through progressive
  distillation.}
  \label{fig:method-overview}
\end{figure*}

\figref{fig:method-overview} provides an overview of \methodname{}, which
decouples capability consolidation from architectural adaptation. Stage~I
consolidates complementary specialist capabilities into a unified
softmax attention Vision Transformer, and Stage~II transfers this representation
to an efficient hybrid attention backbone.

\subsection{Problem Formulation}
\label{sec:method-formulation}

Let $\mathcal{D}$ denote the stage-specific training corpus described in
\secref{sec:data}, and let $\teachers=\{T_k\}_{k=1}^{K}$ denote a set of frozen
visual teachers with complementary capabilities. Given an image $x$, teacher
$T_k$ produces a target representation
$T_k(x)\in\mathbb{R}^{N_k\times d_k}$, where $N_k$ and $d_k$ denote the token
count and feature dimension, respectively ($N_k=1$ for a global embedding).
These targets differ in spatial resolution, feature dimension, and
representational emphasis.

Our objective is to train a hybrid attention encoder $H_{\psi}$, parameterized
by $\psi$, whose representations transfer effectively to recognition, dense
prediction, multimodal understanding, and robotic tasks. We write its output as
\mbox{$H_{\psi}(x)=(\mathbf{H}_{\mathrm{p}}(x),\mathbf{h}_{\mathrm{g}}(x))$}, where
the subscripts $\mathrm{p}$ and $\mathrm{g}$ denote patch and global features,
respectively. Direct multi-teacher distillation jointly optimizes $\psi$ and
the teacher-specific alignment interfaces $\{A_k\}_{k=1}^{K}$ by minimizing
\begin{equation}
  \mathcal{L}_{\mathrm{direct}} =
  \sum_{k=1}^{K}
  \mathcal{L}_k\!\left(A_k(H_{\psi}(x)),\Phi_k(T_k(x))\right),
  \label{eq:direct-distillation}
\end{equation}
where $A_k$ serves as the alignment interface for teacher $T_k$,
$\Phi_k$ normalizes the teacher target, and $\mathcal{L}_k$ matches the resulting
features, as detailed in Stage~I. Spatial alignment of dense features is
implicit in this notation. We use $\mathcal{L}_k$ for teacher-specific
feature-matching losses and stage-specific subscripts for the overall objectives,
written for an image $x$ and aggregated over training minibatches from $\mathcal{D}$.
This objective couples two optimization challenges: reconciling heterogeneous
teacher targets and reproducing them with a different token-mixing architecture.

We address this coupling in two stages. Stage~I jointly trains a softmax attention
capability bridge $B_{\theta}$ and its teacher-specific alignment interfaces
against the frozen specialist ensemble. Stage~II freezes the trained bridge and
uses it as the sole teacher for the hybrid attention encoder $H_{\psi}$.

\subsection{Stage I: Heterogeneous Capability Consolidation}
\label{sec:method-consolidation}

Stage~I learns a shared representation from heterogeneous teachers while
retaining softmax attention. This separates capability consolidation from
adaptation to a different token-mixing operator.

\subsubsection{Complementary Specialist Teachers}
\label{sec:method-teachers}

We select three teachers for their complementary capabilities:
language-aligned semantics, self-supervised spatial structure, and image-level
discriminability.

\begin{itemize}[leftmargin=*,itemsep=2pt,topsep=4pt]
  \item \textbf{QwenViT} (the vision encoder of
  Qwen3-VL-8B~\cite{bai2025qwen3vl}) provides language-aligned patch features
  that transfer decoder-compatible semantics to the bridge.
  \item \textbf{DINOv3 ViT-H+/16} provides spatially precise,
  category-agnostic patch features learned without language supervision. We
  discard its CLS and register tokens.
  \item \textbf{PE-Core-G14} provides a globally discriminative image--text
  representation through its pooled $\ell_2$-normalized embedding, supervising
  the bridge's global representation.
\end{itemize}

All teachers remain frozen and retain their native intensity normalization,
with geometrically consistent views across model inputs. For PE-Core, the input is
resampled to $14g_h\times14g_w$, yielding a patch grid aligned with the
student's $g_h\times g_w$ token layout, where $g_h$ and $g_w$ denote the grid
height and width. Dense targets are spatially aligned with the bridge features
as described in \secref{sec:method-feature-alignment}.

\subsubsection{Capability Bridge}
\label{sec:method-full}
\label{sec:method-feature-alignment}

The capability bridge $B_{\theta}$ is a softmax attention ViT trained from
scratch. Its output is
$B_{\theta}(x)=(\mathbf{F}_{\mathrm{p}}(x),\mathbf{f}_{\mathrm{g}}(x))$, with
dense patch features $\mathbf{F}_{\mathrm{p}}(x)\in\mathbb{R}^{N_{\mathrm{p}}\times d}$
and a global feature $\mathbf{f}_{\mathrm{g}}(x)\in\mathbb{R}^{d}$. Here,
$N_{\mathrm{p}}=g_hg_w$ is the patch token count and $d=1152$ is the shared
hidden dimension of both encoders. The backbone uses $N_{\mathrm{r}}=8$
learnable register tokens, giving an attention sequence length of
$N=N_{\mathrm{p}}+N_{\mathrm{r}}$. A multi-head attention-pooling (MAP) head
derives the global representation from the patch features.

Each teacher has an alignment interface $A_k$ and a target normalization
operator $\Phi_k$. For dense supervision, $A_k$ projects patch features into
the teacher space, with token reordering and spatial resampling as needed to
align prediction and target grids. For global supervision, it projects the
MAP output. The interface includes both the learnable projection and the
required spatial alignment.

Following the feature standardization strategy of
EUPE~\cite{zhu2026eupe}, we normalize each teacher target per channel as
$\Phi_k(T_k(x))=(T_k(x)-\mu_k)/\sigma_k$, where
$\mu_k,\sigma_k\in\mathbb{R}^{d_k}$ are estimated before training and then
held fixed. This reduces differences in target feature statistics. The
alignment interfaces are trained jointly with the bridge in Stage~I and are
not used during Stage~II. Implementation details are provided
in \appref{app:teachers}.

\subsubsection{Progressive Capability Learning}
\label{sec:method-multiteacher-objective}
\label{sec:method-teacher-balancing}
\label{sec:method-stage1-curriculum}

QwenViT and DINOv3 provide dense patch supervision, while PE-Core-G14 provides
global supervision. The matching loss for teacher $T_k$ is
\begin{equation}
  \mathcal{L}_k
  =
  \begin{cases}
    0.9\,\mathcal{L}_{\mathrm{cos}}
      + 0.1\,\mathcal{L}_{\mathrm{sl1}},
      & T_k \text{ is dense}, \\
    \mathcal{L}_{\mathrm{cos}},
      & T_k \text{ is global}.
  \end{cases}
  \label{eq:per-teacher-loss}
\end{equation}
Both terms act on the same aligned prediction and normalized target.
Here, $\mathcal{L}_{\mathrm{cos}}$ is the cosine distance $1-\cos(\cdot,\cdot)$,
and $\mathcal{L}_{\mathrm{sl1}}$ is the smooth-$\ell_1$ loss with
$\beta=10^{-3}$. For dense supervision, both losses are averaged over spatial
positions, with $\mathcal{L}_{\mathrm{sl1}}$ additionally averaged over feature
channels. The cosine term aligns feature directions, while the smooth-$\ell_1$
term also constrains feature magnitudes.

Stage~I jointly optimizes $\theta$ and $\{A_k\}_{k=1}^{K}$ by minimizing
the equally weighted objective
\begin{equation}
  \mathcal{L}_{\mathrm{I}}
  = \sum_{k=1}^{K}
    \mathcal{L}_k\!\left(A_k(B_{\theta}(x)),\Phi_k(T_k(x))\right).
  \label{eq:stage1-objective}
\end{equation}
Training proceeds from fixed-resolution consolidation to native-resolution
refinement, as summarized in \tabref{tab:stage1-curriculum}.

\paragraph{Phase I-A: Fixed-resolution consolidation.}
We first train the bridge on $384\times384$ crops from the curated pretraining
corpus $\mathcal{D}_{\mathrm{pre}}$ (\secref{sec:data}), consolidating teacher
features on a fixed token grid to provide a stable setting to reconcile the heterogeneous teacher targets.

\paragraph{Phase I-B: Native-resolution refinement.}
We continue from the Phase~I-A checkpoint using native aspect ratios, a larger
token budget, and the refinement corpus $\mathcal{D}_{\mathrm{ref}}$.
We lower the learning rate while retaining the same teachers, alignment
interfaces, and objective.

\begin{table}[t]
  \centering
  \small
  \caption{Stage-I training curriculum. Phase~I-A learns the capability bridge
  at fixed resolution, and Phase~I-B refines it using native-resolution inputs.}
  \label{tab:stage1-curriculum}
  \begin{tabularx}{\linewidth}{@{}l
    >{\raggedright\arraybackslash}X
    >{\raggedright\arraybackslash}X@{}}
    \toprule
    & Phase I-A & Phase I-B \\
    \midrule
    Input geometry & Square crop, $384\times384$
      & Native aspect ratio \\
    Token budget & $24\times24=576$ patches
      & $\leq$3920 merged tokens ($\approx$4.0M px) \\
    Peak / final LR & $10^{-3}$ / $10^{-5}$
      & $10^{-4}$ / $10^{-6}$ \\
    Batch / device & 64 & 16 \\
    Training cost & 960 H200 GPU-hours
      & 288 H200 GPU-hours \\
    Training data & $\mathcal{D}_{\mathrm{pre}}$
      & $\mathcal{D}_{\mathrm{ref}}$ \\
    \bottomrule
  \end{tabularx}
\end{table}

\subsection{Stage II: Cross-Architecture Representation Transfer}
\label{sec:method-linearization}

Stage~II transfers the frozen bridge's representation to a hybrid
softmax--linear backbone, following ViT-AdaLA's sequential attention and
feature alignment strategy~\cite{li2026vitadala}.

\subsubsection{Hybrid Attention Student Architecture}
\label{sec:method-gdn}

The final encoder $H_{\psi}$ preserves the bridge's patch embedding, token
layout, feature dimensions, and output interface, while replacing softmax
attention with linear mixers in 18 of the 27 blocks. Compatible parameters are
inherited from the trained bridge. The linear blocks avoid materializing
pairwise attention maps, while the remaining 9 softmax blocks retain exact
global interactions. The complete configuration is provided in
\appref{app:hybrid-student}, and its efficiency--accuracy trade-off is evaluated
in \secref{sec:exp-efficiency}.

\subsubsection{Cross-Architecture Objectives}
\label{sec:method-cross-architecture-objective}

We first align each replacement linear mixer with its frozen softmax
counterpart. Both receive the same normalized block input, with the frozen
softmax path advancing the residual stream. This keeps preceding linear
approximation errors out of the alignment inputs. The objective is
\begin{equation}
  \mathcal{L}_{\mathrm{II}}^{\mathrm{attn}}
  = \frac{1}{|\mathcal{S}_{\mathrm{lin}}|\,Nd}
    \sum_{b\in\mathcal{S}_{\mathrm{lin}}}
    \big\|\mathbf{O}^{\mathrm{lin}}_{b}
          - \mathbf{O}^{\mathrm{soft}}_{b}\big\|^2,
  \label{eq:attention-alignment}
\end{equation}
where $\mathcal{S}_{\mathrm{lin}}$ indexes the replaced attention blocks,
$\mathbf{O}^{\mathrm{soft}}_{b},
\mathbf{O}^{\mathrm{lin}}_{b}\in\mathbb{R}^{N\times d}$ denote the output
token sequences after attention output projection and before residual addition
at block $b$. Here, $N$ includes patch and register tokens. The loss averages squared errors over
selected blocks, tokens, and feature channels.

We then directly match the student's final patch and global features to the
bridge outputs, which have the same dimensions. Using the matching losses in
Eq.~\eqref{eq:per-teacher-loss}, with no teacher-specific alignment or target
standardization, the full-model objective is
\begin{equation}
  \begin{aligned}
    \mathcal{L}_{\mathrm{II}}^{\mathrm{arch}}
      ={}&\mathcal{L}_{\mathrm{cos}}(\mathbf{h}_{\mathrm{g}},\mathbf{f}_{\mathrm{g}}) \\
      &+ 0.9\,\mathcal{L}_{\mathrm{cos}}(\mathbf{H}_{\mathrm{p}},\mathbf{F}_{\mathrm{p}})
       + 0.1\,\mathcal{L}_{\mathrm{sl1}}(\mathbf{H}_{\mathrm{p}},\mathbf{F}_{\mathrm{p}}).
  \end{aligned}
  \label{eq:architecture-distillation}
\end{equation}
Matching only the final outputs leaves intermediate features free to adapt to the new mixers.

\subsubsection{Progressive Architecture-Transfer Curriculum}
\label{sec:method-stage2-curriculum}

The three-phase curriculum progresses from local attention alignment to
full-model distillation and native-resolution refinement
(\tabref{tab:stage2-curriculum}).

\paragraph{Phase II-A: Attention-module alignment.}
We first optimize the replacement linear mixers using
Eq.~\eqref{eq:attention-alignment}, keeping the rest of the transferred
backbone frozen. This initializes the mixers for end-to-end adaptation.

\paragraph{Phase II-B: Fixed-resolution full-model distillation.}
We then unfreeze the complete student and optimize
$\mathcal{L}_{\mathrm{II}}^{\mathrm{arch}}$ at fixed resolution using
$\mathcal{D}_{\mathrm{pre}}$, allowing inherited and replacement components
to co-adapt to the bridge targets.

\paragraph{Phase II-C: Native-resolution refinement.}
Finally, we continue full-model distillation with native aspect ratios and
$\mathcal{D}_{\mathrm{ref}}$, retaining the frozen bridge and
$\mathcal{L}_{\mathrm{II}}^{\mathrm{arch}}$. Optimization settings are reported in
\appref{app:hyperparams}.

\begin{table*}[t]
  \centering
  \small
  \caption{Stage-II architecture-transfer curriculum. The three phases
  progressively expand the trainable parameter set and input geometry while
  retaining the frozen capability bridge as teacher.}
  \label{tab:stage2-curriculum}
  \begin{tabularx}{\textwidth}{@{}l
    >{\raggedright\arraybackslash}X
    >{\raggedright\arraybackslash}X
    >{\raggedright\arraybackslash}X@{}}
    \toprule
    & Phase II-A & Phase II-B & Phase II-C \\
    \midrule
    Objective & $\mathcal{L}_{\mathrm{II}}^{\mathrm{attn}}$
      & $\mathcal{L}_{\mathrm{II}}^{\mathrm{arch}}$
      & $\mathcal{L}_{\mathrm{II}}^{\mathrm{arch}}$ \\
    Trainable parameters & Linear mixers only
      & Complete student & Complete student \\
    Input geometry & Square, $384\times384$
      & Square, $384\times384$ & Native aspect ratio \\
    Token budget & $24\times24=576$ patches
      & $24\times24=576$ patches & $\leq3920$ merged tokens \\
    Peak LR & $10^{-2}$ & $10^{-4}$ & $10^{-4}$ \\
    Batch / device & 32 & 16 & 16 \\
    Training cost & 32 H200 GPU-hours
      & 768 H200 GPU-hours & 288 H200 GPU-hours\\
    Training data & $\mathcal{D}_{\mathrm{pre}}$
      & $\mathcal{D}_{\mathrm{pre}}$ & $\mathcal{D}_{\mathrm{ref}}$ \\
    \bottomrule
  \end{tabularx}
\end{table*}

%% file: main/experiments.tex
\section{Experiments}
\label{sec:experiments}

We evaluate both stages of \methodname{}: the softmax attention capability bridge
\ironvitsoftmax{} and the final hybrid attention encoder \ironvithybrid{}. The two
models share the same macro-architecture and output interfaces, with similar
parameter counts, allowing us to assess how well the consolidated capabilities
are retained after cross-architecture transfer. We compare them with publicly available encoders of similar scale:
DINOv3-L/16~\cite{simeoni2025dinov3},
PE-Core-L/14~\cite{bolya2025perceptionencoder},
SigLIP2-So400M/16-NaFlex~\cite{tschannen2025siglip2},
GenLIP-So/16~\cite{fang2026genlip}, and
C-RADIOv4-SO400M~\cite{ranzinger2026cradiov4}.
Together, these baselines represent self-supervised, contrastive
vision--language, generative language--image, and multi-teacher distillation
paradigms. Their vision encoders contain approximately $0.30$--$0.42$B
parameters. The larger Stage-I teachers are evaluated separately as
non-scale-matched references in \secref{sec:exp-teacher-transfer}.

We use official checkpoints for all encoders and follow the task-specific
preprocessing protocols described in \appref{app:evaluation-protocols}, with model-specific input
normalization. Unless otherwise specified, the visual backbones remain frozen.
Each downstream task consumes the output representation specified in its
corresponding evaluation protocol. In the standard benchmark tables, the best and
second-best results are bolded and underlined, respectively; any task-specific
conventions are stated in the corresponding captions. We evaluate the
generality of the learned representations across five complementary capability
dimensions: visual recognition, image--text retrieval, dense prediction,
multimodal understanding, and robotic learning. Detailed evaluation protocols
are provided in the corresponding subsections.

\subsection{Classification}
\label{sec:exp-cls}

We evaluate image recognition under zero-shot and frozen-feature settings.
Zero-shot classification is evaluated on
ImageNet-1k~\cite{deng2009imagenet} and four distribution-shift variants:
ImageNet-V2~\cite{recht2019imagenetv2},
ImageNet-Sketch~\cite{wang2019imagenetsketch},
ImageNet-Rendition (ImageNet-R)~\cite{hendrycks2021imagenetr}, and
ImageNet-Adversarial (ImageNet-A)~\cite{hendrycks2021imageneta}. For models
with language-aligned
representations, we use the corresponding text tower; our models are coupled
with the PE-Core-G14 text tower through the PE-aligned adapter
$\studadapter{\text{PE}}$, while C-RADIOv4 uses its SigLIP2-g adapter with the
corresponding SigLIP2-giant text tower. We further assess frozen visual
representations on ImageNet-1k using linear probing and $k$-NN classification.
All models follow the same evaluation protocol for each setting. Full
implementation details are provided in
\appref{app:classification-protocol}.

\begin{table}[t]
  \centering
  \footnotesize
  \setlength{\tabcolsep}{3pt}
  \caption{ImageNet recognition accuracy (top-1, \%). Zero-shot results are
  reported on ImageNet-1k (IN-1k), ImageNet-V2 (V2), ImageNet-R (R),
  ImageNet-A (A), and ImageNet-Sketch (Sk); \emph{Avg.} is their unweighted
  mean. Linear probing and $k$-NN are evaluated on ImageNet-1k only. A dash
  indicates that zero-shot evaluation is not applicable.}
  \label{tab:cls}
  \begin{tabular}{@{}l cccccc|c|c@{}}
    \toprule
    & \multicolumn{6}{c|}{Zero-shot} & Linear probe
      & $k$-NN \\
    \cmidrule(lr){2-7} \cmidrule(lr){8-8} \cmidrule(l){9-9}
    Method & IN-1k & V2 & R & A & Sk & Avg. & IN-1k & IN-1k \\
    \midrule
    DINOv3-L/16~\cite{simeoni2025dinov3}
      & -- & -- & -- & -- & -- & --
      & 87.07 & \second{85.42} \\
    PE-Core-L/14~\cite{bolya2025perceptionencoder}
      & \second{83.56} & \best{77.91} & \second{95.31} & \best{90.24} & \second{73.47} & \best{84.10}
      & 87.17 & 85.01 \\
    SigLIP2-So400M-NaFlex~\cite{tschannen2025siglip2}
      & \best{83.81} & \second{77.75} & \best{95.78} & \second{85.56} & \best{75.99} & \second{83.78}
      & \best{87.45} & 85.28 \\
    GenLIP-So/16~\cite{fang2026genlip}
      & -- & -- & -- & -- & -- & --
      & 71.25 & 72.88 \\
    C-RADIOv4-SO400M~\cite{ranzinger2026cradiov4}
      & 81.14 & 75.05 & 94.59 & 80.92 & 68.91 & 80.12
      & 87.33 & 85.04 \\
    \midrule
    \ironvitsoftmax{}
      & 83.27 & 76.91 & 94.13 & 81.99 & 71.18 & 81.50
      & \second{87.35} & \best{85.82} \\
    \ironvithybrid{}
      & 82.88 & 76.20 & 93.48 & 79.89 & 70.22 & 80.53
      & 87.12 & 85.37 \\
    \bottomrule
  \end{tabular}
\end{table}

As shown in \tabref{tab:cls}, \ironvitsoftmax{} yields highly
discriminative frozen representations, ranking first under $k$-NN
classification and second under linear probing. \ironvithybrid{} closely
preserves this capability after architecture transfer. In the zero-shot
setting, PE-Core and SigLIP2 achieve the strongest average performance,
consistent with their direct vision--language pretraining.
Nevertheless, \ironvitsoftmax{} outperforms the comparable multi-teacher
baseline C-RADIOv4 on four of the five datasets and by $1.37$ points on
average, while \ironvithybrid{} remains competitive. These results indicate
that capability consolidation produces strong visual representations with
effective language alignment, and that these properties are largely retained
after transfer to the hybrid attention architecture.

\subsection{Retrieval}
\label{sec:exp-retrieval}

We evaluate cross-modal retrieval on COCO~\cite{lin2014coco} and
Flickr30k~\cite{plummer2015flickr30k}, reporting Recall@$1$ for both
image-to-text and text-to-image retrieval. We use the
same model-specific vision--language interfaces as in zero-shot classification
and exclude models without text-aligned representations. Retrieval-specific
details are provided in \appref{app:retrieval-protocol}.

\begin{table}[t]
  \centering
  \footnotesize
  \caption{Zero-shot image--text retrieval on COCO and Flickr30k (Recall@$1$).
  \emph{Avg.} is the unweighted mean across both datasets and retrieval
  directions. Only models with text-aligned representations are included.}
  \label{tab:retrieval}
  \begin{tabular}{lcccc|c}
    \toprule
    \multirow{2}{*}{Method} & \multicolumn{2}{c}{COCO (R@1)}
      & \multicolumn{2}{c|}{Flickr30k (R@1)}
      & \multirow{2}{*}{Avg.} \\
    \cmidrule(lr){2-3} \cmidrule(lr){4-5}
           & T$\to$I & I$\to$T & T$\to$I & I$\to$T & \\
    \midrule
    PE-Core-L/14~\cite{bolya2025perceptionencoder}
      & \best{56.98} & \best{76.04} & \best{85.74} & \best{95.80}
      & \best{78.64} \\
    SigLIP2-So400M-NaFlex~\cite{tschannen2025siglip2}
      & 56.24 & \second{72.28} & 83.12 & 94.40 & 76.51 \\
    C-RADIOv4-SO400M~\cite{ranzinger2026cradiov4}
      & 56.26 & 71.38 & 83.84 & 94.50 & 76.50 \\
    \midrule
    \ironvitsoftmax{} & \second{56.33} & 71.36 & 84.24 & \second{94.70}
      & \second{76.66} \\
    \ironvithybrid{} & 56.22 & 70.66 & \second{84.92} & 94.10 & 76.48 \\
    \bottomrule
  \end{tabular}
\end{table}

As shown in \tabref{tab:retrieval}, \ironvitsoftmax{} achieves the
second-best overall retrieval performance and slightly outperforms the
comparable multi-teacher baseline C-RADIOv4 on average, demonstrating that
strong cross-modal alignment is retained during capability consolidation.
\ironvithybrid{} largely preserves text-to-image retrieval, but degrades more noticeably in the reverse direction,
suggesting that this retrieval direction is somewhat more sensitive to the representation
changes introduced by softmax-to-hybrid transfer.

\subsection{Dense vision tasks}
\label{sec:exp-dense}

We evaluate the spatial quality of frozen representations on ADE20K semantic
segmentation~\cite{zhou2017ade20k} and NYUv2 monocular depth
estimation~\cite{silberman2012nyuv2}. The two evaluations
follow the DINOv2 linear-probe~\cite{oquab2024dinov2} and TIPS
protocols~\cite{maninis2025tips}, respectively. All encoders use identical
task-specific training and inference settings: ADE20K uses $512^2$ training
and sliding-window crops, whereas NYUv2 retains its native $480\times640$
resolution. This keeps the task-side evaluation settings consistent across
encoders, so performance differences primarily reflect the frozen visual
representations.
Full implementation details are provided in \appref{app:dense-protocol}.

\begin{table}[t]
  \centering
  \footnotesize
  \setlength{\tabcolsep}{5pt}
  \caption{Dense prediction with frozen visual encoders. We report mIoU on
  ADE20K semantic segmentation and RMSE on NYUv2 depth estimation.}
  \label{tab:dense}
  \begin{tabular}{@{}l cc@{}}
    \toprule
    Method & ADE20K mIoU $\uparrow$ & NYUv2 RMSE $\downarrow$ \\
    \midrule
    DINOv3-L/16~\cite{simeoni2025dinov3}
      & 52.86 & \second{0.320} \\
    PE-Core-L/14~\cite{bolya2025perceptionencoder}
      & 40.57 & 0.611 \\
    SigLIP2-So400M-NaFlex~\cite{tschannen2025siglip2}
      & 44.48 & 0.459 \\
    GenLIP-So/16~\cite{fang2026genlip}
      & 45.59 & 0.484 \\
    C-RADIOv4-SO400M~\cite{ranzinger2026cradiov4}
      & \best{54.10} & \best{0.295} \\
    \midrule
    \ironvitsoftmax{} & \second{52.97} & 0.339 \\
    \ironvithybrid{} & 52.6 & 0.343 \\
    \bottomrule
  \end{tabular}
\end{table}

As shown in \tabref{tab:dense}, \ironvitsoftmax{} ranks second on ADE20K
segmentation and remains competitive with the spatially specialized DINOv3 on
NYUv2 depth, while outperforming all language-pretrained baselines on both
tasks. \ironvithybrid{} closely matches its segmentation performance and incurs
only a modest degradation in depth estimation, indicating that dense spatial
representations are largely preserved after architecture transfer. Together
with the classification and retrieval results, these findings show that
\methodname{} remains competitive across both semantic and spatial
evaluations, despite the different strengths of language-aligned and
dense-prediction baselines.

\subsection{VLM visual encoder}
\label{sec:exp-vlm}

We evaluate each encoder as the vision tower in a controlled
LLaVA-NeXT~\cite{liu2024llavanext} pipeline with a
Qwen2.5-7B-Instruct decoder~\cite{qwen2025qwen25}. The connector, training data,
and optimization schedule are held fixed across models to enable a controlled
comparison of the visual representations. Following LLaVA-NeXT, the vision encoder is frozen
during connector alignment and unfrozen for joint optimization during the
second-stage instruction tuning. Evaluation spans $22$ benchmarks covering OCR and
document understanding, knowledge and diagram reasoning, vision-centric
perception, and general multimodal understanding. Full training
and evaluation details are provided in \appref{app:vlm-protocol}.

\begin{table}[!t]
  \centering
  \footnotesize
  \setlength{\tabcolsep}{4pt}
  \caption{Multimodal understanding with each encoder as the vision tower in a
  controlled LLaVA-NeXT setup. Italicized rows report unweighted family means,
  and the final row averages all $22$ benchmarks. Entries are accuracy unless
  noted; the raw MME Perception score is divided by $20$ when computing means.}
  \label{tab:vlm}
  \begin{tabular}{@{}l ccccc cc@{}}
    \toprule
    & DINOv3-L/16 & PE-Core-L/14 & SigLIP2 & GenLIP & C-RADIOv4 & \multicolumn{2}{c}{\methodname{}} \\
    \cmidrule(l){7-8}
    Benchmark & \cite{simeoni2025dinov3} & \cite{bolya2025perceptionencoder} & \cite{tschannen2025siglip2} & \cite{fang2026genlip} & \cite{ranzinger2026cradiov4} & Softmax & Hybrid \\
    \midrule
    \emph{OCR and document understanding}   & 41.90 & 57.43 & 57.94 & \second{61.34} & 57.73 & \best{61.52} & 61.14 \\
    \cmidrule(lr){2-8}
    \quad TextVQA~\cite{singh2019textvqa}   & 47.25 & 66.83 & 68.16 & \best{70.55} & 66.40 & \second{68.89} & 67.86 \\
    \quad DocVQA~\cite{mathew2021docvqa}    & 51.46 & 72.93 & 71.38 & \best{77.30} & 70.99 & \second{76.99} & 76.34 \\
    \quad OCRBench~\cite{liu2024ocrbench} \tiny(total acc.)
                                          & 36.60 & 54.80 & 57.40 & 62.50 & 57.60 & \best{64.20} & \second{62.90} \\
    \quad ChartQA~\cite{masry2022chartqa}   & 57.72 & 72.64 & 72.32 & 74.64 & 73.44 & \second{75.68} & \best{75.80} \\
    \quad OCRBench-v2~\cite{fu2025ocrbenchv2} \tiny(total acc.)
                                          & 16.47 & 19.95 & 20.42 & 21.72 & 20.21 & \second{21.85} & \best{22.82} \\
    \addlinespace[2pt]
    \midrule
    \addlinespace[1pt]
    \emph{Knowledge and diagram reasoning}  & 78.04 & 79.66 & 80.86 & 79.86 & 80.40 & \best{81.29} & \second{80.96} \\
    \cmidrule(lr){2-8}
    \quad ScienceQA~\cite{lu2022scienceqa}  & 79.63 & \second{82.08} & \best{82.53} & 80.57 & 81.40 & 82.06 & 81.99 \\
    \quad AI2D~\cite{kembhavi2016ai2d}      & 76.46 & 77.23 & 79.18 & 79.15 & 79.40 & \best{80.51} & \second{79.92} \\
    \addlinespace[2pt]
    \midrule
    \addlinespace[1pt]
    \emph{Vision-centric perception}        & 60.29 & 62.84 & \best{66.37} & 63.95 & 64.01 & \second{64.48} & 64.12 \\
    \cmidrule(lr){2-8}
    \quad RealWorldQA~\cite{xai2024realworldqa}
                                          & 63.14 & 63.27 & \best{65.23} & 63.79 & 64.44 & \second{64.84} & 63.66 \\
    \quad POPE~\cite{li2023pope} \tiny(F1)  & \second{87.72} & 87.38 & 87.19 & 87.54 & \best{88.10} & 86.43 & 87.24 \\
    \quad MMVP~\cite{tong2024mmvp}          & 46.67 & 52.67 & \second{54.00} & 52.00 & \best{54.67} & 52.00 & 50.00 \\
    \quad CV-Bench-2D~\cite{tong2024cambrian} \tiny(Overall)
                                          & 63.38 & 68.00 & \second{68.34} & 67.36 & \best{68.83} & 66.88 & 65.82 \\
    \quad CV-Bench-3D~\cite{tong2024cambrian} \tiny(Overall)
                                          & 61.00 & \second{68.83} & 63.33 & 68.42 & \best{69.50} & 66.08 & 66.00 \\
    \quad WhatsUp~\cite{kamath2023whatsup}  & \best{69.23} & 65.74 & 65.56 & 66.65 & 65.05 & \second{68.26} & 67.93 \\
    \quad HallusionBench~\cite{guan2024hallusionbench}
                                          & 58.46 & 58.90 & 60.94 & 60.50 & \second{61.38} & \best{63.24} & 61.29 \\
    \quad BLINK~\cite{fu2024blink}          & 42.71 & \best{44.13} & 42.19 & \second{43.82} & 41.50 & 42.35 & 42.82 \\
    \quad CountBenchQA~\cite{paiss2023countbench}
                                          & 50.31 & 56.67 & \best{90.55} & 65.50 & 62.63 & 70.23 & \second{72.28} \\
    \addlinespace[2pt]
    \midrule
    \addlinespace[1pt]
    \emph{General multimodal understanding} & 64.46 & 67.08 & 67.97 & 66.92 & 67.76 & \best{68.55} & \second{68.08} \\
    \cmidrule(lr){2-8}
    \quad GQA~\cite{hudson2019gqa}          & 64.10 & \best{65.65} & 65.20 & 64.96 & \second{65.41} & 64.99 & 65.04 \\
    \quad MME~\cite{fu2023mme} \tiny(Perception)
                                          & 1492.04 & 1636.25 & \second{1640.09} & 1609.74 & 1586.16 & \best{1650.76} & 1630.15 \\
    \quad MMStar~\cite{chen2024mmstar}      & 49.47 & 50.60 & 52.80 & 52.27 & 54.07 & \best{56.27} & \second{55.20} \\
    \quad SEEDBench-IMG~\cite{li2023seedbench}
                                          & 72.37 & 74.51 & \second{74.89} & 74.00 & \best{75.00} & 74.18 & 74.27 \\
    \quad MMBench-dev-EN~\cite{liu2024mmbench}
                                          & 78.61 & 82.77 & 84.62 & 81.54 & 84.20 & \best{85.31} & \second{84.87} \\
    \quad MMMU~\cite{yue2024mmmu}           & 47.62 & 47.14 & \second{48.29} & \second{48.29} & \best{48.57} & 48.00 & 47.62 \\
    \midrule
    \textbf{Average} (all $22$)        & 58.86 & 64.30 & \second{66.21} & 65.62 & 65.10 & \best{66.44} & 66.05 \\
    \bottomrule
  \end{tabular}
\end{table}

As shown in \tabref{tab:vlm}, \ironvitsoftmax{} achieves the best overall
average and leads three of the four capability families: OCR and document
understanding, knowledge and diagram reasoning, and general multimodal
understanding. It also ranks second on vision-centric perception, indicating
that the consolidated representation transfers broadly rather than relying on
a single benchmark category. \ironvithybrid{} remains competitive after
architecture transfer, ranking second in both knowledge and diagram reasoning
and general multimodal understanding, with a $0.39$-point reduction in the
overall average. Together with its strong dense-prediction performance, these
results demonstrate that \methodname{} combines multimodal semantic reasoning
with spatial perception, while the hybrid attention model retains most of this
generality under a more efficient architecture.

\subsection{Robotic learning}
\label{sec:exp-robotics}

We evaluate frozen visual representations for embodied control on
CortexBench~\cite{majumdar2023vc1}, using two dexterous-manipulation tasks from
Adroit~\cite{rajeswaran2018adroit} (\texttt{pen}, \texttt{relocate}) and five
tabletop tasks from Meta-World~\cite{yu2020metaworld} (\texttt{assembly},
\texttt{bin-picking}, \texttt{button-press-topdown}, \texttt{drawer-open}, and
\texttt{hammer}). Following the Theia protocol~\cite{shang2024theia}, visual
features are coupled with a lightweight behavior-cloning policy while the
encoder remains frozen. All models share the same policy, demonstrations, and
evaluation protocol; implementation details are provided in
\appref{app:robotics-protocol}.

\begin{table}[!t]
  \centering
  \footnotesize
  \setlength{\tabcolsep}{4pt}
  \caption{CortexBench manipulation success rate (\%). Italicized rows average
  the tasks within each suite, and the final row averages all seven tasks.
  Subscripts denote standard deviations across seeds.}
  \label{tab:robotics}
  \begin{tabular}{@{}l ccccc cc@{}}
    \toprule
    & DINOv3-L/16 & PE-Core-L/14 & SigLIP2 & GenLIP & C-RADIOv4 & \multicolumn{2}{c}{\methodname{}} \\
    \cmidrule(l){7-8}
    Benchmark & \cite{simeoni2025dinov3} & \cite{bolya2025perceptionencoder} & \cite{tschannen2025siglip2} & \cite{fang2026genlip} & \cite{ranzinger2026cradiov4} & Softmax & Hybrid \\
    \midrule
    \emph{Adroit}~\cite{rajeswaran2018adroit}
      & 58.00 & 63.35 & 58.00 & 64.00
      & \second{70.67} & \best{71.30} & 68.65 \\
    \cmidrule(lr){2-8}
    \quad pen
      & 73.33\pmstd{6.11} & 70.7\pmstd{6.1} & 73.33\pmstd{8.33} & 73.33\pmstd{8.33}
      & \second{74.67}\pmstd{2.31} & \best{77.3}\pmstd{2.3} & \best{77.3}\pmstd{4.6} \\
    \quad relocate
      & 42.67\pmstd{9.24} & 56.0\pmstd{4.0} & 42.67\pmstd{9.24} & 54.67\pmstd{6.11}
      & \best{66.67}\pmstd{6.11} & \second{65.3}\pmstd{6.1} & 60.0\pmstd{4.0} \\
    \addlinespace[2pt]
    \midrule
    \addlinespace[1pt]
    \emph{Meta-World}~\cite{yu2020metaworld}
      & 91.20 & 75.74 & 73.07 & 83.73
      & \second{91.46} & \best{92.02} & 91.18 \\
    \cmidrule(lr){2-8}
    \quad assembly
      & 94.67\pmstd{9.24} & 82.7\pmstd{10.1} & 77.33\pmstd{18.90} & 93.33\pmstd{8.33}
      & \best{97.33}\pmstd{4.62} & 96.0\pmstd{6.9} & \second{97.3}\pmstd{4.6} \\
    \quad bin-picking
      & 77.33\pmstd{8.33} & 76.0\pmstd{6.9} & 66.67\pmstd{6.11} & 73.33\pmstd{2.31}
      & 73.33\pmstd{12.22} & \second{82.7}\pmstd{4.6} & \best{85.3}\pmstd{12.9} \\
    \quad button-press-topdown
      & \second{88.00}\pmstd{4.00} & 52.0\pmstd{4.0} & 64.00\pmstd{8.00} & 65.33\pmstd{6.11}
      & \best{89.33}\pmstd{4.62} & 86.7\pmstd{4.6} & 81.3\pmstd{8.3} \\
    \quad drawer-open
      & \best{100.00}\pmstd{0.00} & \best{100.00} & \best{100.00}\pmstd{0.00} & \best{100.00}\pmstd{0.00}
      & \best{100.00}\pmstd{0.00} & \best{100.0}\pmstd{0.0} & \best{100.0}\pmstd{0.0} \\
    \quad hammer
      & \second{96.00}\pmstd{4.00} & 68.0\pmstd{12.0} & 57.33\pmstd{8.33} & 86.67\pmstd{9.24}
      & \best{97.33}\pmstd{2.31} & 94.7\pmstd{6.1} & 92.0\pmstd{4.0} \\
    \midrule
    \textbf{Average}
      & 81.71 & 72.20 & 68.76 & 78.09
      & \second{85.52} & \best{86.10} & 84.74 \\
    \bottomrule
  \end{tabular}
\end{table}

\ironvitsoftmax{} achieves the highest overall success rate and leads the
aggregate performance on both Adroit and Meta-World, demonstrating consistent
transfer across dexterous and tabletop manipulation rather than gains confined
to a single task family. This balanced performance shows that multi-teacher
consolidation preserves visual features that generalize effectively to
closed-loop control. Following architecture transfer, \ironvithybrid{} retains
$98.4\%$ of the overall success rate of \ironvitsoftmax{}, while attaining the
best result on \texttt{bin-picking} and matching it on \texttt{pen}. The small
aggregate gap between the two variants indicates that the action-relevant
structure of the consolidated representation is largely preserved by the more
efficient hybrid attention architecture.

\subsection{Capability retention from specialist teachers}
\label{sec:exp-teacher-transfer}

To assess Stage-I capability consolidation, we compare \ironvitsoftmax{} with
each specialist teacher in its respective capability domain: language-aligned
recognition and retrieval for PE-Core, spatial representation learning for
DINOv3, and multimodal understanding for QwenViT. Frozen-feature recognition
and dense prediction provide complementary measures of general visual quality.
Together, these evaluations quantify how effectively the language, spatial,
and multimodal capabilities of the three teachers are consolidated within a
single encoder.

\paragraph{Multimodal capability.}
We substitute \ironvitsoftmax{} for QwenViT in the controlled LLaVA-NeXT setup
of \secref{sec:exp-vlm}.
\tabref{tab:teacher-vlm} reports capability-family and overall results; the
complete benchmark-level comparison is provided in
\appref{app:vlm-detailed}.

\begin{table}[!t]
  \centering
  \footnotesize
  \setlength{\tabcolsep}{4pt}
  \caption{Multimodal downstream performance relative to the QwenViT teacher
  under the controlled setup of \secref{sec:exp-vlm}. Family and
  overall means follow \tabref{tab:vlm}.}
  \label{tab:teacher-vlm}
  \begin{tabular}{@{}l cc@{}}
    \toprule
    Benchmark & QwenViT~\cite{bai2025qwen3vl} & \ironvitsoftmax{} \\
    \midrule
    OCR and document understanding
      & 63.65 & 61.52 \\
    Knowledge and diagram reasoning
      & 80.41 & \best{81.29} \\
    Vision-centric perception
      & 62.94 & \best{64.48} \\
    General multimodal understanding
      & 68.20 & \best{68.55} \\
    \midrule
    \textbf{Average}
      & 66.12 & \best{66.44} \\
    \bottomrule
  \end{tabular}
\end{table}

Under the same downstream adaptation protocol, \ironvitsoftmax{} matches the
overall performance of QwenViT and performs slightly better in knowledge and
diagram reasoning, vision-centric perception, and general multimodal
understanding. This indicates that the decoder-facing capability learned from
QwenViT is largely retained alongside the complementary semantic and spatial
supervision from the other teachers. The remaining gap in
OCR and document understanding identifies fine-grained text recognition as the
least preserved aspect of the teacher's multimodal capability.

\begin{table*}[t]
  \footnotesize
  \setlength{\tabcolsep}{4pt}
  \caption{Comparison with the Stage-I specialist teachers across recognition,
  retrieval, and dense prediction. \emph{Shift} denotes the mean over the four
  ImageNet distribution-shift benchmarks.}
  \label{tab:teacher-single}
  \begin{tabular*}{\textwidth}{@{\extracolsep{\fill}}lcccccccccc@{}}
    \toprule
    & \multicolumn{4}{c}{Recognition}
      & \multicolumn{4}{c}{Retrieval (R@1)}
      & \multicolumn{2}{c}{Dense prediction} \\
    \cmidrule(lr){2-5} \cmidrule(lr){6-9} \cmidrule(l){10-11}
    & \multicolumn{2}{c}{Zero-shot} & $k$-NN & Linear probe
      & \multicolumn{2}{c}{COCO} & \multicolumn{2}{c}{Flickr30k}
      & ADE20K & NYUv2 \\
    \cmidrule(lr){2-3} \cmidrule(lr){6-7} \cmidrule(lr){8-9}
    Encoder & IN-1k & Shift & IN-1k & IN-1k
      & T$\to$I & I$\to$T & T$\to$I & I$\to$T
      & mIoU $\uparrow$ & RMSE $\downarrow$ \\
    \midrule
    PE-Core-G14~\cite{bolya2025perceptionencoder}
      & \best{85.24} & \best{86.16} & \best{86.68} & 86.84
      & \best{57.48} & \best{74.98} & \best{85.16} & \best{95.10}
      & 39.38 & 0.602 \\
    DINOv3-H+/16~\cite{simeoni2025dinov3}
      & -- & -- & 85.55 & \best{87.59} & -- & -- & -- & --
      & 52.83 & \best{0.327} \\
    QwenViT~\cite{bai2025qwen3vl}
      & -- & -- & 80.22 & 78.28 & -- & -- & -- & --
      & 33.81 & 0.503 \\
    \midrule
    \ironvitsoftmax{}
      & 83.27 & 81.05 & 85.82 & 87.35
      & 56.33 & 71.36 & 84.24 & 94.70
      & \best{52.97} & 0.339 \\
    \bottomrule
  \end{tabular*}
\end{table*}

\paragraph{Language-aligned capability.}
Compared with PE-Core-G14, \ironvitsoftmax{} largely preserves zero-shot
retrieval performance, nearly matching the teacher on Flickr30k and remaining
competitive on COCO. The larger gap in zero-shot classification, especially
under distribution shift, suggests that class-level alignment is more
sensitive to consolidation than instance-level cross-modal matching. This
asymmetry may partly reflect the distillation objective: the student
matches PE-Core's visual representations without being jointly optimized with
its text tower, which may alter the cross-modal geometry required for zero-shot
prediction. Overall, the results demonstrate effective
retention of language-aligned capability while incorporating supervision from
the other specialist teachers.

\paragraph{Frozen and spatial representations.}
On frozen-feature recognition, \ironvitsoftmax{} remains close to the strongest
teacher under both $k$-NN and linear probing, while clearly outperforming
QwenViT. Its spatial features are similarly well preserved: the
student slightly exceeds DINOv3-H+/16 on ADE20K segmentation and incurs only a
small increase in NYUv2 depth error. These results indicate that consolidation
maintains the discriminative and spatial structure of the specialist features,
rather than trading them for language or multimodal alignment.

\subsection{Efficiency at high resolution}
\label{sec:exp-efficiency}

\begin{figure*}[t]
  \centering
  \begin{subfigure}[t]{0.49\textwidth}
    \centering
    \includegraphics[width=\linewidth,trim=0 750 0 0,clip]{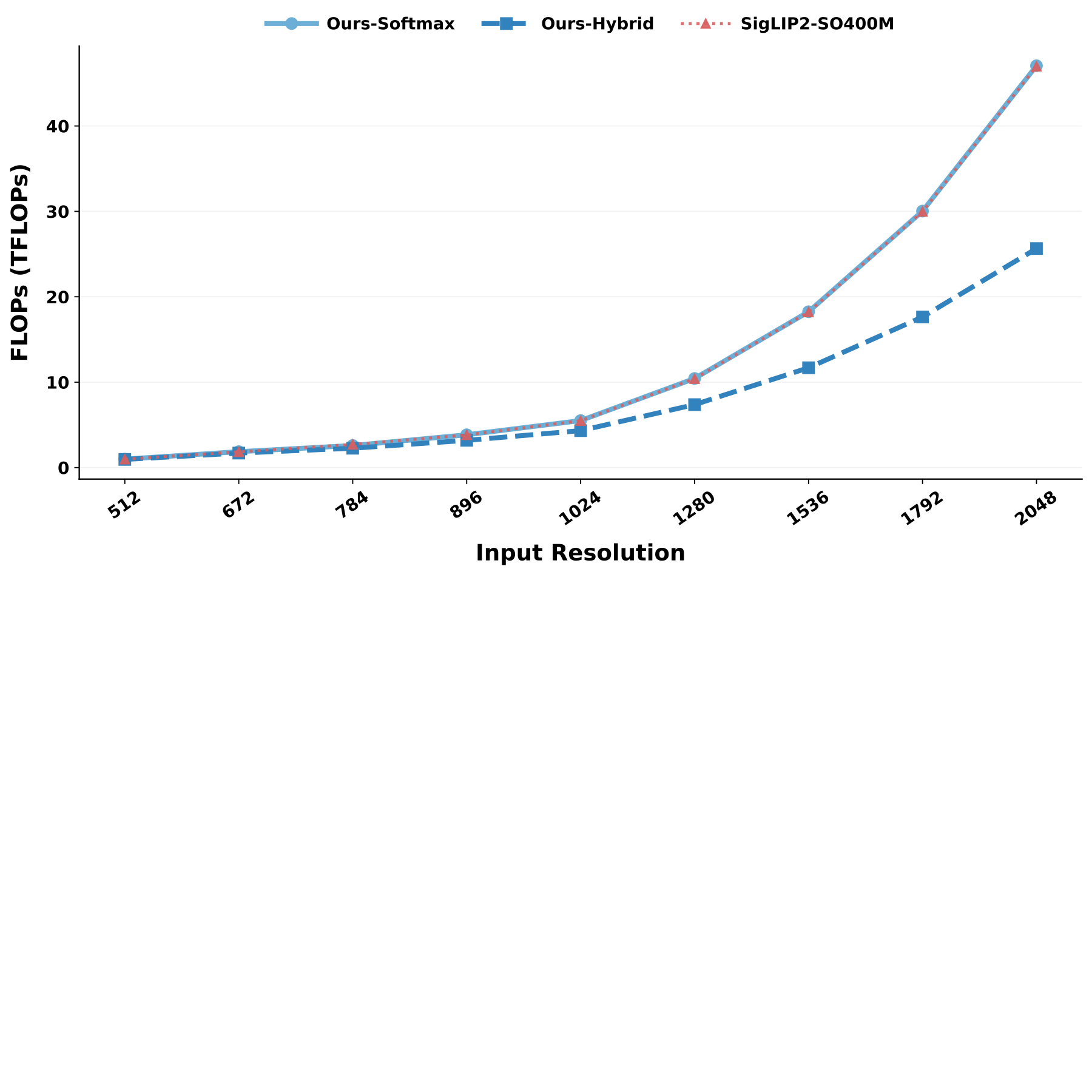}
    \caption{Forward FLOPs per image.}
    \label{fig:efficiency-flops}
  \end{subfigure}
  \hfill
  \begin{subfigure}[t]{0.49\textwidth}
    \centering
    \includegraphics[width=\linewidth,trim=0 750 0 0,clip]{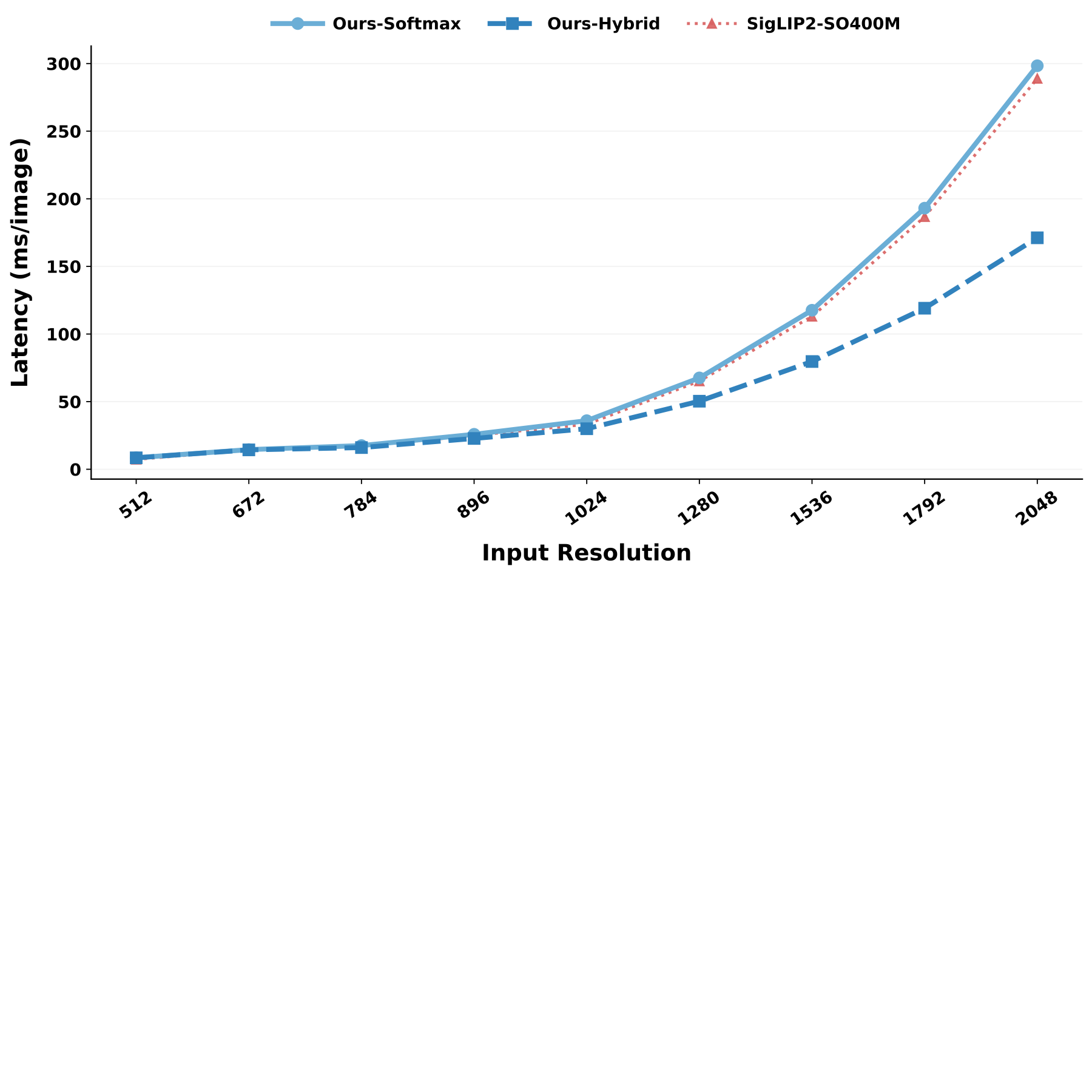}
    \caption{Batch-$1$ inference latency.}
    \label{fig:efficiency-latency}
  \end{subfigure}
  \caption{Computational scaling with input resolution under a shared
  evaluation setup.}
  \label{fig:efficiency}
\end{figure*}

We compare the forward FLOPs and inference latency of \ironvithybrid{},
its softmax counterpart \ironvitsoftmax{}, and SigLIP2-So400M-NaFlex across
input resolutions from $512$ to $2048$ pixels per side
(\figref{fig:efficiency}). SigLIP2 provides an external baseline of comparable
model size with the same patch size.
We report forward FLOPs per image using two FLOPs per multiply--accumulate,
excluding
normalization, elementwise activations, softmax exponentiation, and RoPE.

Latency is measured at batch size $1$ in bfloat16 on the same GPU, using
CUDA Graph execution for all three models. We report the median CUDA-event
time after warm-up, with inputs allocated on the GPU to exclude host-to-device
transfer. Both IronViT variants use a fixed-resolution inference path with
cached positional and sequence metadata, fused QK normalization and RoPE,
and an equivalent 2D patch embedding for static images. These optimizations
reduce redundant computation, synchronization, and memory traffic while
preserving the model's mathematical function. 
For SigLIP2, the redundant mask is omitted for unpadded inputs to enable the optimized
attention backend.

\figref{fig:efficiency} shows a growing efficiency advantage for
\ironvithybrid{} as resolution increases, with lower FLOPs and latency than
both softmax baselines and the largest gains at high resolution.
At $2048\times2048$, it requires approximately $45\%$ fewer FLOPs and achieves
about $1.75\times$ speedup relative to \ironvitsoftmax{}. This trend is
consistent with the increasing contribution of quadratic attention as the
token count grows: replacing $18$ of $27$ softmax blocks reduces this cost,
while shared projection and feed-forward operations limit the gains at lower
resolutions. The slightly smaller latency improvement relative to the FLOPs
reduction reflects the influence of implementation and hardware efficiency.
Combined with the preceding capability results, these findings demonstrate
that the hybrid architecture preserves broad visual competence while
substantially reducing the cost of high-resolution inference.

\subsection{Ablation studies}
\label{sec:exp-ablation}

We examine three design choices used throughout the main experiments: the
construction of the training corpus, the two-stage factorization of
multi-teacher distillation, and the coarse-to-fine curriculum within each
stage. Unless otherwise stated, variants keep the backbone scale, teacher
ensemble, optimization budget, and evaluation protocol fixed.

\subsubsection{Data curation}
\label{sec:exp-abl-data}

We evaluate the data pipeline at two stages of training. First, we study how
progressive corpus curation affects representation learning during
the fixed-resolution consolidation stage. We then evaluate the contribution of semantic enrichment
introduced during the subsequent native-resolution refinement stage.

\begin{figure}[t]
    \centering
    \includegraphics[width=\linewidth]{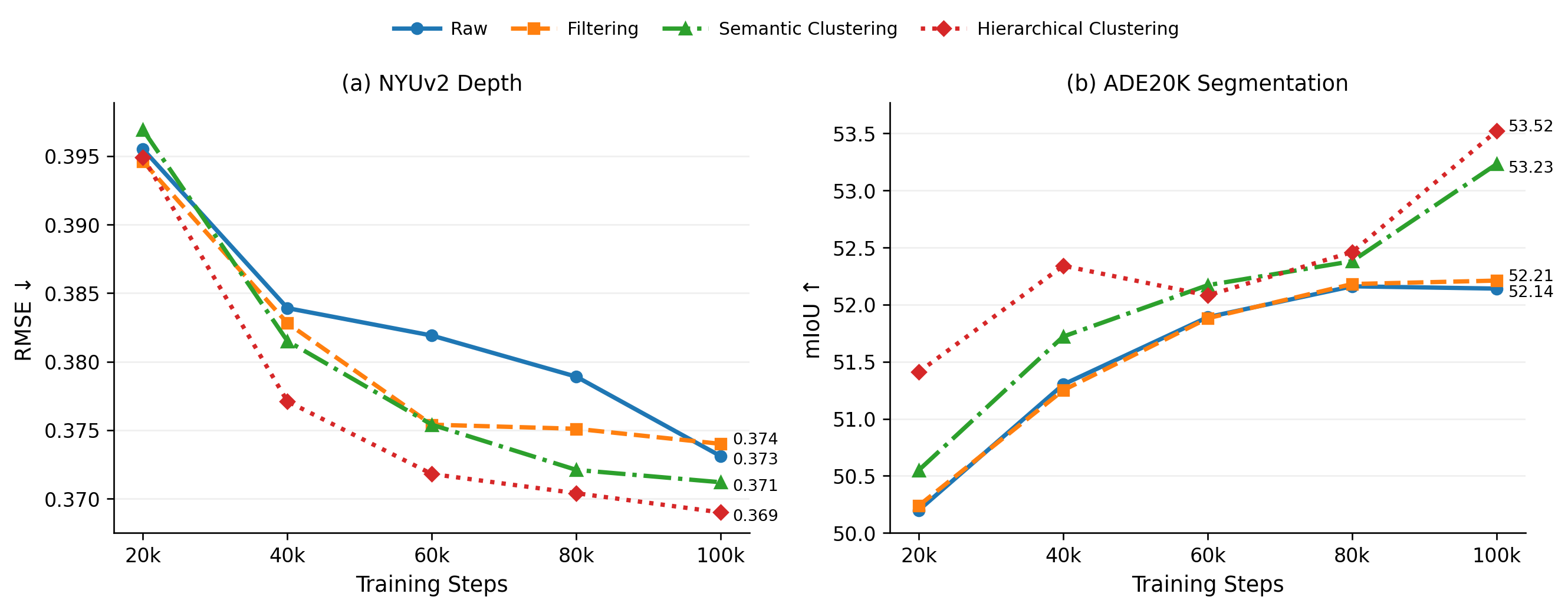}
    \caption{
    \textbf{Training dynamics under progressively curated data.}
    All variants are trained for 100K steps under the same multi-teacher
    configuration and evaluated every 20K steps. We report
    (a) NYUv2 monocular depth estimation (RMSE, lower is better) and
    (b) ADE20K semantic segmentation (mIoU, higher is better).
    }
    \label{fig:abl-data}
\end{figure}

\paragraph{Pretraining corpus curation.}
We conduct a cumulative ablation under a fixed multi-teacher training setup.
Starting from the raw image pool, we progressively apply filtering and
deduplication, semantic clustering with in-cluster pruning, and
hierarchical balanced sampling. All other training settings are kept
identical. As shown in \tabref{tab:abl-data}, progressively curated data consistently
improves representation quality across ImageNet robustness and dense prediction. The gain 
is most pronounced on dense tasks, which are particularly sensitive to the 
diversity and spatial coverage of the pretraining corpus. \figref{fig:abl-data}
further shows that semantic clustering and hierarchical balancing provide the
largest gains on ADE20K and NYUv2, with the improvement emerging early during 
training rather than only at the final checkpoint. These results suggest that
reducing semantic redundancy and balancing visual concept coverage are 
important contributors to the benefit of the curation pipeline.

\begin{table*}[t]
  \centering
  \scriptsize
  \setlength{\tabcolsep}{4.5pt}
  \caption{
  \text{Cumulative ablation of the data curation pipeline.}
  All variants use the same student architecture, multi-teacher setup, and
  training configuration, and differ only in the training corpus.
  We report final-checkpoint performance after 100K training steps on
  ImageNet robustness benchmarks and dense prediction tasks.
  }
  \label{tab:abl-data}

  \begin{tabular}{@{}l cccc cc@{}}
    \toprule
    & \multicolumn{4}{c}{ImageNet Robustness}
    & \multicolumn{2}{c}{Dense Prediction} \\
    \cmidrule(lr){2-5}
    \cmidrule(lr){6-7}

    Data
    & IN-V2 $\uparrow$
    & IN-Sketch $\uparrow$
    & IN-A $\uparrow$
    & IN-R $\uparrow$
    & ADE20K mIoU $\uparrow$
    & NYUv2 RMSE $\downarrow$ \\
    \midrule

    Raw (85M)
      & 77.42 & 63.60 & 73.39 & 86.96
      & 52.14 & 0.373 \\

    Filtering \& Deduplication (79M)
      & 77.28 & 63.56 & 73.48 & 86.88
      & 52.21 & 0.374 \\

    Semantic Clustering (64M)
      & 77.87 & 63.60 & \best{74.55} & 87.48
      & 53.23 & 0.371 \\

    Hierarchical Balanced Sampling (40M)
      & \best{78.06} & \best{63.89} & 73.53 & \best{87.96}
      & \best{53.52} & \best{0.369} \\

    \bottomrule
  \end{tabular}
\end{table*}

\paragraph{Semantic enrichment during refinement.}
We next evaluate the semantic enrichment applied during the refinement stage.
Unlike the preceding ablation, which modifies the composition of the
pretraining corpus, this experiment examines whether introducing
semantically enriched data during refinement further improves the learned
representation. We compare the checkpoint immediately before refinement
against the resulting checkpoint after semantic-enriched refinement, while
keeping the backbone architecture and evaluation protocols unchanged.

As shown in \tabref{tab:abl-semantic-enrichment}, semantic enrichment leads
to clear improvements across both recognition and dense prediction tasks.
These results indicate that semantic enrichment during refinement
complements the corpus-level curation performed during pretraining,
further improving both semantic robustness and dense spatial representations.

\begin{table*}[t]
  \centering
  \scriptsize
  \setlength{\tabcolsep}{4.5pt}
  \caption{
  \textbf{Ablation of semantic enrichment during the refinement stage.}
  We compare the model before refinement with the model after applying
  semantic-enriched refinement. The backbone architecture and evaluation
  protocols are kept unchanged. }
  \label{tab:abl-semantic-enrichment}

  \begin{tabular}{@{}l l cccc cc@{}}
    \toprule
    & 
    & \multicolumn{4}{c}{ImageNet Robustness}
    & \multicolumn{2}{c}{Dense Prediction} \\
    \cmidrule(lr){3-6}
    \cmidrule(lr){7-8}

    Variant
    & Refinement Data
    & IN-V2 $\uparrow$
    & IN-Sketch $\uparrow$
    & IN-A $\uparrow$
    & IN-R $\uparrow$
    & ADE20K mIoU $\uparrow$
    & NYUv2 RMSE $\downarrow$ \\
    \midrule

    Before refinement
      & 44.9M
      & 73.63
      & 67.74
      & 80.72
      & 90.30
      & 53.50
      & 0.365 \\

    + Semantic enrichment
      & 46.4M
      & \best{74.65}
      & \best{68.71}
      & \best{81.16}
      & \best{92.37}
      & \best{53.97}
      & \best{0.349} \\

    \midrule

    $\Delta$
      & +1.5M
      & +1.02
      & +0.97
      & +0.44
      & +2.07
      & +0.47
      & $-0.016$ \\

    \bottomrule
  \end{tabular}
\end{table*}

\subsubsection{Two-stage distillation}
\label{sec:exp-abl-distill}
\label{sec:exp-abl-twostage}
\label{sec:exp-cl}

We next test whether the capability bridge is necessary. The direct baseline
optimises the hybrid student against the full teacher ensemble in a single
multi-teacher distillation run, following Eq.~\eqref{eq:direct-distillation}.
This baseline asks the hybrid model to solve two difficult problems
simultaneously: reconciling heterogeneous teacher targets and adapting those
targets to a different token-mixing architecture. In contrast, our two-stage
procedure first consolidates teacher capabilities into the softmax attention
bridge, and only then transfers the consolidated representation to the hybrid
student. \tabref{tab:abl-two-stage} compares the two strategies across
recognition, dense prediction, and VLM evaluation. For compactness, it reports
the mean over the five zero-shot ImageNet benchmarks, ImageNet-1k linear-probe
and $k$-NN accuracy, ADE20K mIoU, NYUv2 RMSE, and the mean over the $22$ VLM
benchmarks. Complete benchmark-level results are provided in
\appref{app:vlm-detailed}.

\begin{table*}[t]
  \centering
  \footnotesize
  \setlength{\tabcolsep}{3pt}
  \caption{Two-stage distillation ablation.}
  \label{tab:abl-two-stage}
  \begin{tabular}{@{}l ccc cc c@{}}
    \toprule
    & \multicolumn{3}{c}{Classification}
      & \multicolumn{2}{c}{Dense prediction}
      & VLM \\
    \cmidrule(lr){2-4} \cmidrule(lr){5-6} \cmidrule(l){7-7}
    Strategy & ZS $\uparrow$ & LP $\uparrow$ & $k$-NN $\uparrow$
      & Seg. $\uparrow$ & Depth $\downarrow$ & Avg. $\uparrow$ \\
    \midrule
    Direct multi-teacher
      & 75.11 & 85.37 & 83.39 & 52.27 & 1.046 & 58.15 \\
    Two-stage (ours)
      & \best{80.53} & \best{87.12} & \best{85.37} & \best{52.6}
      & \best{0.343} & \best{66.05} \\
    \bottomrule
  \end{tabular}
\end{table*}

The two-stage strategy consistently outperforms direct multi-teacher
distillation across all evaluation dimensions. The improvements are most
pronounced in zero-shot classification, depth estimation, and VLM evaluation,
with the VLM average increasing by $7.90$ points and NYUv2 RMSE decreasing from
$1.046$ to $0.343$. Gains in linear probing, $k$-NN, and segmentation further
show that the benefit extends to both global discrimination and dense spatial
representations. These results are consistent with the proposed factorization:
under this comparison, introducing a softmax attention bridge before
cross-architecture transfer yields stronger final representations than direct
multi-teacher distillation into the hybrid student.

\subsubsection{Multi-phase training curriculum}
\label{sec:exp-abl-curriculum}

Finally, we evaluate the repeated training phases used inside each stage. The
primary focus is Stage~I, where Phase~I-A learns a stable fixed-resolution
multi-teacher representation and Phase~I-B refines it under native aspect
ratios and a larger token budget. This comparison tests whether the second
training pass merely extends optimization or provides a distinct benefit by
exposing the bridge to higher-resolution and variable-geometry inputs.
\tabref{tab:abl-stage1-curriculum} reports the two Stage-I checkpoints across
classification, dense prediction, and VLM evaluation families; complete
benchmark-level VLM results are provided in \appref{app:vlm-detailed}.

\begin{table*}[t]
  \centering
  \footnotesize
  \setlength{\tabcolsep}{3pt}
  \caption{Stage-I curriculum ablation across classification, dense prediction,
  and VLM evaluation families.}
  \label{tab:abl-stage1-curriculum}
  \begin{tabular}{@{}l c cc ccccc@{}}
    \toprule
    & Classification & \multicolumn{2}{c}{Dense prediction}
      & \multicolumn{5}{c}{VLM} \\
    \cmidrule(lr){2-2} \cmidrule(lr){3-4} \cmidrule(l){5-9}
    Stage-I checkpoint & IN-1k LP & Seg. & Depth
      & OCR & Know. & Percep. & General & Avg. \\
    \midrule
    Phase I-A
      & 86.56 & 52.51 & 0.384 & 59.34 & 80.5 & 62.97 & 68.23 & 65.17 \\
    ${}+$ Phase I-B
      & \best{87.35} & \best{52.97} & \best{0.339} & \best{61.52} & \best{81.29} & \best{64.48} & \best{68.55} & \best{66.44} \\
    \bottomrule
  \end{tabular}
\end{table*}

Adding Phase~I-B improves every reported evaluation. Beyond gains in linear
probing and segmentation, NYUv2 RMSE decreases from $0.384$ to $0.339$,
consistent with stronger spatial transfer after the complete Phase~I-B
refinement. The VLM
improvements are largest in OCR and vision-centric perception, raising the
overall average by $1.27$ points while also improving the knowledge and general
understanding categories. This consistent pattern shows that native-resolution
refinement complements fixed-resolution consolidation, strengthening geometric
and visually grounded capabilities without compromising global recognition or
multimodal reasoning.

%% file: main/conclusion.tex
\section{Conclusion}
\label{sec:conclusion}

We presented \methodname{}, a framework for learning efficient generalist
visual representations by separating capability consolidation from
architectural adaptation. Supported by curated data, a softmax attention
capability bridge unifies complementary teacher representations and enables
their transfer to a hybrid attention encoder. Experiments demonstrate broad
downstream transfer, substantial efficiency gains at high resolution, and
consistent improvements over direct multi-teacher distillation. Together,
these findings suggest that establishing a shared representation before
constraining its computation provides a practical path toward efficient
generalist vision. The retained softmax blocks still impose quadratic cost,
and transfer to video and real-world robotic deployment remains to be
established.

Looking forward, \methodname{} provides a foundation for visual encoders that
can grow in both capability and domain coverage. Mixture-of-experts
architectures offer a natural route to scaling capability consolidation:
specialized representation pathways and adaptive routing could accommodate an
expanding set of heterogeneous teachers while reducing interference among
their inductive biases. Beyond architectural scaling, continual learning could
transform distillation from a one-off compression procedure into an evolving
learning process. By continually absorbing deployment data, task supervision,
and embodied feedback, the resulting encoder could adapt to the distinctive
visual distributions encountered in autonomous driving and robotics,
including specialized sensing geometries such as fisheye cameras. This could
enable the student to acquire capabilities beyond those explicitly transferred
from a fixed teacher ensemble, progressing from capability inheritance toward
capability acquisition.

%% file: main/appendix.tex
\section{Additional Details}
\label{app:details}

This appendix documents the data curation settings, model architectures, training configurations, and downstream evaluation protocols used in our experiments.
It also provides benchmark-level results for the VLM comparisons and ablation studies.

\subsection{Data curation details}
\label{app:data-curation}

\tabref{tab:curation-config} summarizes the data curation settings.
The pipeline and corpus composition are described in \secref{sec:data}
and \secref{sec:data-recipe}.
Unless otherwise specified, similarity is measured by cosine similarity
between $L_2$-normalized embeddings.

\begin{table*}[t]
  \centering
  \footnotesize
  \setlength{\tabcolsep}{5pt}
  \renewcommand{\arraystretch}{1.15}
  \caption{
  \textbf{Implementation settings for the data curation pipeline.}
  }
  \label{tab:curation-config}

  \begin{tabular}{@{}
    p{0.19\textwidth}
    p{0.24\textwidth}
    p{0.51\textwidth}
    @{}}
    \toprule
    Module & Setting & Configuration \\
    \midrule

    Perceptual deduplication
      & Hashing
      & 64-bit DCT-based perceptual hash
        (hash size 8; high-frequency factor 4) \\

      & Image preprocessing
      & Lanczos downsampling to $\max(H,W)\leq4{,}096$;
        aspect ratio preserved \\

      & Duplicate criterion
      & Hamming distance $=0$;
        highest-resolution representative retained \\

    \midrule

    Semantic clustering
      & Image encoder
      & Frozen CLIP ViT-L/14 and ViT-H/14 encoders \\

      & $k$-means
      & Spherical $k$-means;
        $K=20{,}000$;
        50 iterations;
        $\leq256$ training samples per centroid \\

      & Semantic pruning
      & Descending order of centroid distance;
        comparison with all preceding samples;
        $\varepsilon=0.15$ \\

    \midrule

    \multirow[t]{4}{=}{\raggedright Hierarchical balanced sampling}
      & Hierarchy
      & $(K_1,K_2,K_3)=(20{,}000,5{,}000,1{,}000)$,
        from finest to coarsest \\

      & Level-1 fitting
      & $5\times10^6$ embeddings sampled uniformly without replacement;
        50 iterations;
        seed $=0$ \\

      & Resampling
      & 10 rounds per level;
        $(15,5,2)$ centroid-nearest representatives per cluster
        at Levels~1--3 \\

      & Final selection
      & Capacity-constrained recursive balanced allocation;
        uniform sampling without replacement within leaf clusters \\

    \midrule

    Semantic enrichment
      & Image anchors
      & Centroid-nearest seed images;
        CLIP ViT-L/14;
        50 target-domain $k$-means iterations \\

      & Text anchors
      & 5--8 prompts per domain, 59 total;
        corresponding CLIP text encoder \\

    \bottomrule
  \end{tabular}
\end{table*}

\subsection{Training details}

\subsubsection{Training hyperparameters}
\label{app:hyperparams}

Teachers remain frozen and use \texttt{bfloat16} precision; the student uses
\texttt{bfloat16} mixed precision with FP32 master weights.
Two-dimensional weight matrices are optimized with
MuonPlus~\cite{zhang2026muonplus}.
Biases, normalization parameters, and other one-dimensional parameters use
AdamW. All parameter groups use linear
learning-rate warm-up followed by cosine decay. Teacher and student inputs
are geometrically aligned, with model-specific normalization.

In Stage~I, updates are skipped when the combined gradient norm exceeds
$5.0$ after the first $2{,}000$ steps. In Stage~II, weight decay is set to
$0.05$ for Phase~II-A and $0.01$ for Phases~II-B and II-C, with 2-step
gradient accumulation in the latter two phases. Other phase-specific
settings are summarized in Tables~\ref{tab:stage1-curriculum}
and~\ref{tab:stage2-curriculum}.

\subsubsection{Per-teacher alignment}
\label{app:teachers}

The learnable projection within each teacher-specific alignment interface $A_k$
is a $d\!\rightarrow\!1536\!\rightarrow\!d_k$ MLP, where $d=1152$, with
bias-free linear layers separated by LayerNorm and GELU. Teacher and student
patch tokens are aligned in row-major order for dense supervision, accounting for Qwen3-VL's
native $2\times2$ blocked layout. When their grids differ, features are
bicubically resampled to a common grid whose height and width are the respective
maxima of the two grids. Per-channel teacher statistics
$(\mu_k,\sigma_k)$ are estimated over 500 data-loader iterations before
training and remain fixed thereafter.

\subsubsection{Backbone architectures}
\label{app:hybrid-student}

\begin{figure*}[t]
  \centering
  \includegraphics[width=\textwidth]{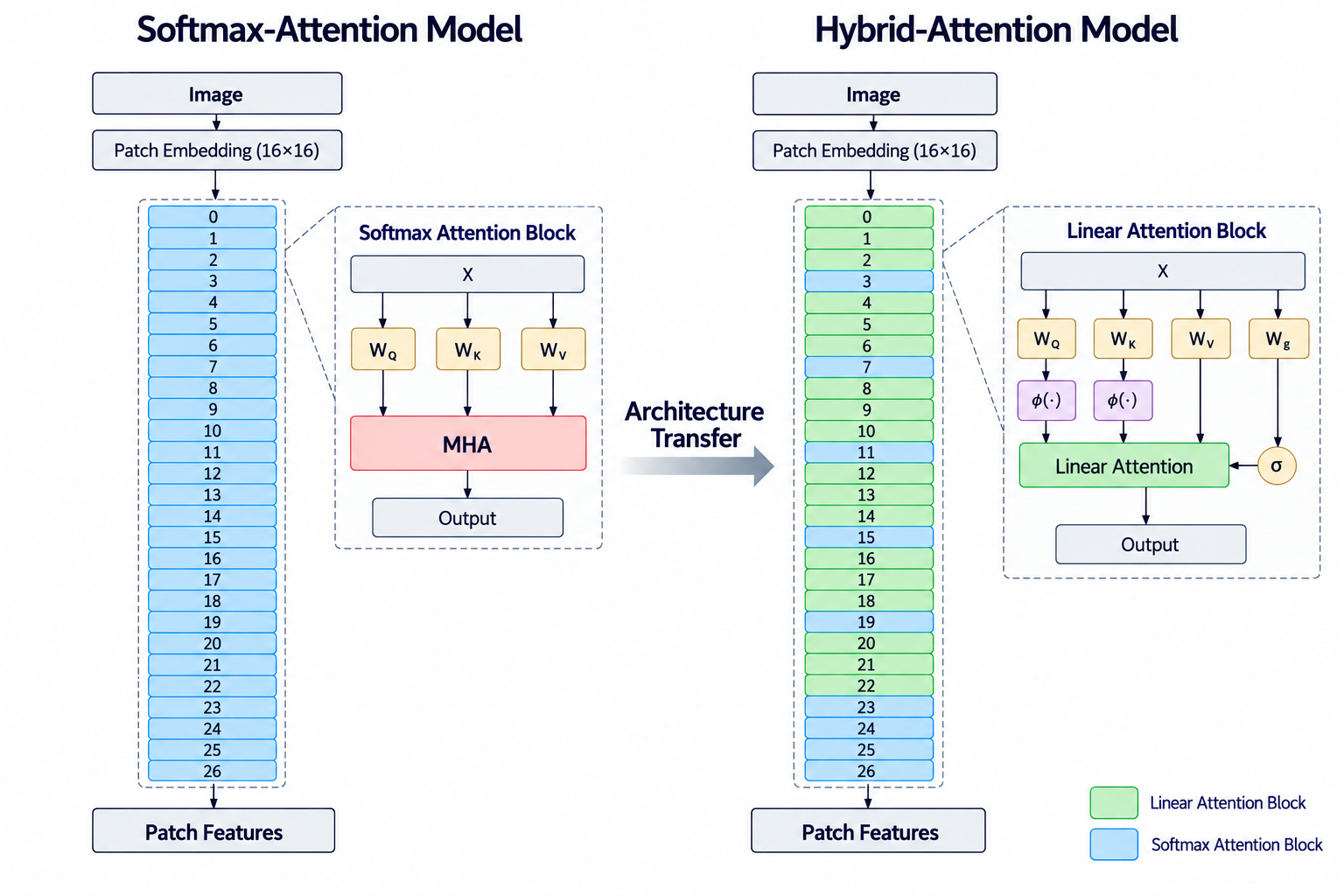}
  \caption{Architectural overview of the softmax and hybrid attention models.}
  \label{fig:hybrid-architecture}
\end{figure*}

\paragraph{Architecture.}
The capability bridge and hybrid model share a 27-block pre-LayerNorm ViT
with hidden width $d=1152$, MLP width 4,304, 16 attention heads,
$16\times16$ patches, and $N_{\mathrm{r}}=8$ register tokens. Both use
interpolated absolute positional embeddings and 2D RoPE.

A single-query MAP head without query--key normalization pools the
layer-normalized patch features. Register tokens participate in backbone
attention but are excluded from pooling.

The bridge and hybrid backbones
contain 416M and 440M parameters, respectively, including linear-attention
gates but excluding the MAP head and teacher-specific adapters.

\paragraph{Attention.}
The bridge uses softmax attention in all 27 blocks. The hybrid model retains
it at zero-based indices
$\{3,7,11,15,19,23,24,25,26\}$ and uses linear attention in the remaining
18 blocks. For a single head, let $\mathbf{Q}$, $\mathbf{K}$, and $\mathbf{V}$
denote the query, key, and value matrices. The softmax attention output for
token $i$ is
\begin{equation}
\mathbf{O}^{\mathrm{soft}}_{i}=\sum_{j=1}^{N}
 \frac{\exp(\mathbf{Q}_{i}\mathbf{K}_{j}^{\top})}
 {\sum_{k=1}^{N}\exp(\mathbf{Q}_{i}\mathbf{K}_{k}^{\top})}
 \mathbf{V}_{j}.
 \label{eq:backbone-softmax}
\end{equation}
Linear-attention blocks use the feature map
$\phi(\mathbf{x})=1+\operatorname{ELU}(\mathbf{x})$:
\begin{equation}
 \mathbf{O}^{\mathrm{lin}}_{i}
 =\sigma(\mathbf{X}_{i}\mathbf{W}_{g})\odot
 \frac{\phi(\mathbf{Q}_{i})
 \sum_{j=1}^{N}\phi(\mathbf{K}_{j})^{\top}\mathbf{V}_{j}}
 {\phi(\mathbf{Q}_{i})\sum_{k=1}^{N}\phi(\mathbf{K}_{k})^{\top}}.
 \label{eq:backbone-linear}
\end{equation}
Here, $N=N_{\mathrm{p}}+N_{\mathrm{r}}$ counts patch and register tokens,
$\mathbf{X}_i$ is the input representation of token $i$, and $\mathbf{W}_g$ is the gate projection;
$\sigma$ and $\odot$ denote sigmoid and element-wise multiplication.
The gate projection is zero-initialized, giving an initial gate value of
$0.5$, and the gated output is passed through the output projection.
Query--key normalization and 2D RoPE are omitted from the equations for brevity.

\subsection{Evaluation protocols and results}
\label{app:evaluation-protocols}

\subsubsection{Data curation ablation}
\label{app:data-ablation}

Each corpus variant is used to train an independent capability bridge for
$100$k steps under the Phase~I-A configuration
(\tabref{tab:stage1-curriculum}), with checkpoints evaluated every $20$k
steps. Only the training corpus varies; all other training and evaluation
settings are held fixed. Additional training and alignment details are
provided in \appref{app:hyperparams} and \appref{app:teachers}.

\subsubsection{Classification}
\label{app:classification-protocol}

All classification experiments use frozen visual encoders with model-specific
input normalization. Global image features are obtained from the CLS token,
mean-pooled patch tokens, or a MAP head, depending on the architecture.
\methodname{} uses the MAP output.

\paragraph{Zero-shot classification.}
For \methodname{}, the MAP features are mapped to the PE-specific alignment
head and transformed back to the PE-Core-G14 embedding space by inverting
the distillation-time target standardization. Images are bicubically resized
to $384\times384$ without center cropping. Class prototypes are obtained by averaging the
$L_2$-normalized text embeddings of 80 English CLIP
prompts~\cite{radford2021clip}. Predictions use cosine similarity to these
prototypes, restricted to the corresponding 200 classes for ImageNet-A
and ImageNet-R.

\paragraph{Linear probing.}
Images are bicubically resized to a shorter side of 438 pixels while
preserving the aspect ratio, then center-cropped to $384\times384$.
We train a linear classifier for 20 epochs
with cross-entropy and Nesterov SGD (batch size 1,024, peak learning rate 0.1),
without weight decay or label smoothing. A 2-epoch warm-up is
followed by cosine learning-rate decay. We report the highest validation
top-1 accuracy.

\paragraph{$k$-NN classification.}
We report 20-NN top-1 accuracy using the full ImageNet-1k training set as
the reference set. Neighbors are ranked by cosine similarity between
$L_2$-normalized features, with class votes weighted by $\exp(s/0.07)$,
where $s$ denotes cosine similarity.

\subsubsection{Image--text retrieval}
\label{app:retrieval-protocol}

We evaluate on the Karpathy test splits of COCO (5,000 images)
and Flickr30k (1,000 images). Before tokenization, captions are
lowercased, stripped of selected punctuation, and truncated to 50 words.
Evaluation uses the same text encoders, image adapters, and image preprocessing
as zero-shot classification, without retrieval-specific training.
\methodname{} pairs its PE-aligned MAP features with the PE-Core-G14 text encoder.

We rank the full test-set gallery by cosine similarity between
$L_2$-normalized image and text embeddings and report Recall@1 in both
retrieval directions (\tabref{tab:retrieval}).

\subsubsection{Dense prediction}
\label{app:dense-protocol}

Visual encoders remain frozen with native input normalization; only task heads
are trained. Dense features are reordered into raster order when needed.

\paragraph{ADE20K.}
We use the standard ADE20K split with 150 semantic categories.
Patch features are concatenated with a
spatially broadcast global feature, then processed by BatchNorm and a
$1\times1$ convolution. The resulting logits are bilinearly upsampled to the
input resolution for cross-entropy supervision, with unlabeled pixels ignored.

Training uses random scaling ($s\sim\mathcal{U}(0.5,2.0)$),
$512\times512$ random crops,
horizontal flipping, and photometric distortion. We train the head for
$40$k iterations using AdamW (batch size 16, peak learning rate $10^{-3}$,
weight decay $10^{-4}$), with $1.5$k warm-up iterations followed by linear
learning-rate decay.

At inference, images are resized to a shorter side of 512 pixels while
preserving the aspect ratio. Sliding-window prediction uses $512\times512$
crops with a $341\times341$ stride, averaging logits in overlapping regions.
We report the highest validation mIoU.

\paragraph{NYUv2.}
We use the standard BTS/Eigen split at the native $480\times640$ resolution.
Patch features are
concatenated with their broadcast spatial mean, bilinearly upsampled by
$4\times$, and projected to 256 depth-bin logits using a $1\times1$
convolution. After interpolation to the image resolution, depth is computed
as the softmax-weighted mean of bin centers uniformly spaced between
$10^{-3}$ and 10 meters.

We train the head for $50$k iterations using AdamW (batch size 8, peak
learning rate $10^{-4}$, weight decay 0.01), with $16$k linear warm-up
iterations followed by cosine learning-rate decay.
The loss combines scale-invariant log loss ($\lambda=0.15$; log-RMSE during
the first 100 iterations) with a depth-gradient term weighted by 0.5.
Training augmentation includes random rotation, horizontal flipping, and
color jitter.

We average predictions from original and horizontally flipped images and
evaluate valid depths in $(10^{-3},10)$ meters within the Eigen crop.
We report RMSE for the checkpoint with the highest $\delta_1$ accuracy on
the test set.

\subsubsection{VLM}
\label{app:vlm-protocol}

Following LLaVA-NeXT, we first train a randomly initialized MLP connector on
558,000 image--caption pairs while freezing both the vision encoder and
language model (learning rate $10^{-3}$, global batch size 256). We then
unfreeze the vision encoder for 1 epoch of joint instruction tuning on
739,000 multimodal instruction examples. Learning rates are $2\times10^{-5}$
for the language model and $2\times10^{-6}$ for the vision encoder, with a
global batch size of 128, 3\% warm-up, and no weight decay.

Images are processed using AnyRes multi-crop, and only patch tokens are
passed to the connector. Global batch sizes are matched across models.

\subsubsection{Robotic learning}
\label{app:robotics-protocol}

Adroit and Meta-World use 100 and 25 expert demonstrations per task,
respectively. Frames are rendered at $256\times256$ and resized to
$224\times224$.
Following Theia~\cite{shang2024theia}, three convolution--BatchNorm layers
map the spatial features to a 256-dimensional representation. The policy is
trained by behavior cloning for 100 epochs using Adam (learning rate
$10^{-3}$, batch size 256) and an MSE loss. We evaluate 25 episodes every
5 epochs and report success rates across 3 training seeds.

\subsubsection{Detailed VLM results}
\label{app:vlm-detailed}

\tabref{tab:vlm-detailed} reports benchmark-level VLM results for the QwenViT
baseline~\cite{bai2025qwen3vl}, the Phase I-A and I-B checkpoints, and the
hybrid models obtained through direct and two-stage distillation.
\begin{table}[!t]
  \centering
  \scriptsize
  \setlength{\tabcolsep}{3.5pt}
  \renewcommand{\arraystretch}{1.06}
  \caption{Detailed VLM results. Family scores are unweighted means, and the
  overall score averages all $22$ benchmarks, with MME Perception divided by
  $20$ before averaging (as in \tabref{tab:vlm}). Entries are accuracy unless
  noted; the best overall score is bolded.}
  \label{tab:vlm-detailed}
  \begin{tabularx}{\textwidth}{@{}>{\raggedright\arraybackslash}X c c c c c@{}}
    \toprule
    & \multicolumn{1}{c}{Teacher baseline}
    & \multicolumn{2}{c}{Stage-I curriculum}
    & \multicolumn{2}{c}{Transfer strategy} \\
    \cmidrule(lr){2-2} \cmidrule(lr){3-4} \cmidrule(l){5-6}
    \raisebox{.5\normalbaselineskip}{Benchmark}
      & \raisebox{.5\normalbaselineskip}{QwenViT}
      & \raisebox{.5\normalbaselineskip}{Phase I-A}
      & \shortstack{\ironvitsoftmax{}\\(Phase I-B)}
      & \raisebox{.5\normalbaselineskip}{Direct}
      & \raisebox{.5\normalbaselineskip}{\ironvithybrid{}} \\
    \midrule
    \emph{OCR and document understanding}
      & 63.65 & 59.34 & 61.52 & 43.77 & 61.14 \\
    \cmidrule(lr){2-6}
    \quad TextVQA~\cite{singh2019textvqa}
      & 71.04 & 66.44 & 68.89 & 54.75 & 67.86 \\
    \quad DocVQA~\cite{mathew2021docvqa}
      & 79.69 & 74.22 & 76.99 & 49.16 & 76.34 \\
    \quad OCRBench~\cite{liu2024ocrbench}
      & 66.10 & 59.80 & 64.20 & 46.20 & 62.90 \\
    \quad ChartQA~\cite{masry2022chartqa}
      & 77.64 & 75.00 & 75.68 & 53.36 & 75.80 \\
    \quad OCRBench-v2~\cite{fu2025ocrbenchv2}
      & 23.77 & 21.24 & 21.85 & 15.39 & 22.82 \\
    \addlinespace[2pt]
    \midrule
    \addlinespace[1pt]
    \emph{Knowledge and diagram reasoning}
      & 80.41 & 80.50 & 81.29 & 76.44 & 80.96 \\
    \cmidrule(lr){2-6}
    \quad ScienceQA~\cite{lu2022scienceqa}
      & 81.42 & 81.49 & 82.06 & 79.79 & 81.99 \\
    \quad AI2D~\cite{kembhavi2016ai2d}
      & 79.40 & 79.50 & 80.51 & 73.09 & 79.92 \\
    \addlinespace[2pt]
    \midrule
    \addlinespace[1pt]
    \emph{Vision-centric perception}
      & 62.94 & 62.97 & 64.48 & 58.56 & 64.12 \\
    \cmidrule(lr){2-6}
    \quad RealWorldQA~\cite{xai2024realworldqa}
      & 62.35 & 63.14 & 64.84 & 61.05 & 63.66 \\
    \quad POPE~\cite{li2023pope} \tiny(F1)
      & 86.39 & 87.55 & 86.43 & 87.03 & 87.24 \\
    \quad MMVP~\cite{tong2024mmvp}
      & 44.67 & 48.00 & 52.00 & 38.67 & 50.00 \\
    \quad CV-Bench-2D~\cite{tong2024cambrian}
      & 66.01 & 63.79 & 66.88 & 60.90 & 65.82 \\
    \quad CV-Bench-3D~\cite{tong2024cambrian}
      & 63.92 & 68.58 & 66.08 & 64.17 & 66.00 \\
    \quad WhatsUp~\cite{kamath2023whatsup}
      & 68.50 & 64.74 & 68.26 & 63.71 & 67.93 \\
    \quad HallusionBench~\cite{guan2024hallusionbench}
      & 62.18 & 61.82 & 63.24 & 53.76 & 61.29 \\
    \quad BLINK~\cite{fu2024blink}
      & 40.77 & 41.77 & 42.35 & 41.93 & 42.82 \\
    \quad CountBenchQA~\cite{paiss2023countbench}
      & 71.66 & 67.35 & 70.23 & 55.85 & 72.28 \\
    \addlinespace[2pt]
    \midrule
    \addlinespace[1pt]
    \emph{General multimodal understanding}
      & 68.20 & 68.23 & 68.55 & 63.41 & 68.08 \\
    \cmidrule(lr){2-6}
    \quad GQA~\cite{hudson2019gqa}
      & 64.33 & 64.79 & 64.99 & 62.86 & 65.04 \\
    \quad MME~\cite{fu2023mme} \tiny(Perception)
      & 1655.89 & 1634.50 & 1650.76 & 1537.32 & 1630.15 \\
    \quad MMStar~\cite{chen2024mmstar}
      & 55.73 & 55.13 & 56.27 & 44.40 & 55.20 \\
    \quad SEEDBench-IMG~\cite{li2023seedbench}
      & 73.62 & 74.27 & 74.18 & 70.75 & 74.27 \\
    \quad MMBench-dev-EN~\cite{liu2024mmbench}
      & 84.71 & 85.15 & 85.31 & 79.79 & 84.87 \\
    \quad MMMU~\cite{yue2024mmmu}
      & 48.00 & 48.29 & 48.00 & 45.81 & 47.62 \\
    \midrule
    \textbf{Average} (all $22$)
      & 66.12 & 65.17 & \best{66.44} & 58.15 & 66.05 \\
    \bottomrule
  \end{tabularx}
\end{table}